\newcommand{\isforarchiv}{}
\RequirePackage{fix-cm}
\ifx\isforarchiv\undefiened
\documentclass[twocolumn]{svjour3}          
\else
\documentclass[twocolumn]{svjour3_arxiv}          
\fi
\smartqed  
\usepackage{tabularx}
\usepackage[latin1]{inputenc}
\usepackage[cmex10]{amsmath}
\usepackage{amsfonts}
\usepackage{amssymb}
\usepackage{amsbsy} 
\usepackage{hyperref}
\usepackage{comment}
\usepackage[scriptsize,tight]{subfigure}
\usepackage[]{units}
\usepackage{bigstrut}
\usepackage{wrapfig}
\usepackage[pdftex]{graphicx}
\graphicspath{{./figures/}}
\DeclareGraphicsExtensions{.pdf}
\usepackage[]{todonotes}

\renewenvironment{remark}[1][Remark]{\begin{trivlist} \item[\hskip \labelsep {\bfseries #1}]}{\end{trivlist}}

\journalname{Autonomous Robots}
\begin{document}
\title{ Momentum Control with Hierarchical Inverse Dynamics on a Torque-Controlled Humanoid
%
}

\author{Alexander Herzog$^{1}$ \and Nicholas Rotella$^{2}$ \and Sean Mason$^{2}$  \and Felix Grimminger$^{1}$ \and Stefan Schaal$^{1,2}$ \and Ludovic Righetti$^{1,2}$}

\institute{ \at
           $^1$Autonomous Motion Dept., Max-Planck Inst. Intelligent Systems,
           72076 T\"ubingen, Germany\\
           $^2$Computational Learning and Motor Control Lab, University of
           Southern California, Los Angeles, CA 90089, USA \\
           \and
           Alexander Herzog \at
           Autonomous Motion Dept., Max-Planck Inst. Intelligent Systems,
           72076 T\"ubingen, Germany\\ 
           \email{aherzog@tuebingen.mpg.de} 
}

\ifx\isforarchiv\undefiened
\date{Received: date / Accepted: date}
\else
\date{}
\fi

\maketitle

\begin{abstract}
Hierarchical inverse dynamics based on cascades of quadratic
programs have been proposed for the control of legged robots. They
have important benefits but to the best of our knowledge have never
been implemented on a torque controlled humanoid where model
inaccuracies, sensor noise and real-time computation requirements
can be problematic. Using a reformulation of existing algorithms, we
propose a simplification of the problem that allows to achieve
real-time control. Momentum-based control is integrated in the task
hierarchy and a LQR design approach is used to compute the desired
associated closed-loop behavior and improve performance. Extensive
experiments on various balancing and tracking tasks show very robust
performance in the face of unknown disturbances, even when the
humanoid is standing on one foot. Our results demonstrate that
hierarchical inverse dynamics together with momentum control can be
efficiently used for feedback control under real robot conditions.
\end{abstract}

\section{Introduction}
\footnote{Part of the material presented in this paper has been
  presented at the 2014 IEEE/RSJ International Conference on
  Intelligent Robots and Systems}We expect autonomous legged robots to
perform complex tasks in persistent interaction with an uncertain and
changing environment (e.g. in a disaster relief scenario). Therefore,
we need to design algorithms that can generate precise but compliant
motions while optimizing the interactions with the environment. In
this context, the choice of a control strategy for legged robots is of
primary importance as it can drastically improve performance in the
face of unexpected disturbances and therefore open the way for agile
robots, whether they are locomoting or performing manipulation tasks.

Robots with torque control
capabilities~\cite{Boaventura:2012tc,Hutter:2012uu}, including
humanoids~\cite{Cheng:2008tw,Moro:2013,Ott:2011uj}, are becoming
increasingly available and torque control algorithms are therefore
necessary to fully exploit their capabilities. Indeed, such algorithms
often offer high performance for motion control while guaranteeing a
certain level of compliance
\cite{Boaventura:2012tc,Kalakrishnan:2011dy,Saab:2013vr,Salini:2011bw}.
In addition, they also allow for the direct control of contact
interactions with the environment
\cite{Hutter:2012uu,Righetti:2013tt,righetti11b}, which is required
during operation in dynamic and uncertain environments. Recent
contributions have also demonstrated the relevance of torque control
approaches for humanoid robots
\cite{Hyon:2007jya,Ott:2011uj,Stephens:2010vu}. We can essentially
distinguish two control approaches.

\textit{Passivity-based} approaches on humanoids were originally
proposed in \cite{Hyon:2007jya} and recently extended in
\cite{Ott:2011uj}. They compute admissible contact forces and control
commands under quasi-static assumptions. The great advantage of such
approaches is that they do not require a precise dynamic model of the
robot. Moreover, robustness is generically guaranteed due to the
passivity property of the controllers. However, the quasi-static
assumption can be a limitation for dynamic motions.

On the other hand, controllers based on the \textit{full dynamic
  model} of the robot have also been successfully implemented on
legged robots
\cite{Hutter:2012uu,Righetti:2013tt,Stephens:2010vu,Mason:2014,Vaillant:2014cl}.
These methods essentially perform a form of inverse dynamics. The
advantage of such approaches is that they are in theory well suited
for very dynamic motions. However, sensor noise (particularly in the
velocities), limited torque bandwidth and the need for a precise
dynamic model make them more challenging to implement. Moreover, it is
generally required to simplify the optimization process to meet time
requirements of fast control loops (typically \unit[1]{kHz} on modern
torque controlled robots). Although there are many contributions
showing the potential of such approaches in simulation
\cite{Faraji:2014,Feng:2014,Kuindersma:2014,Salini:2011bw},
evaluations on real robots are still rare due to the lack of torque
controlled humanoid platforms and the complexity in conducting such
experiments.

Extensions of the inverse dynamics approach have been proposed where
it is also possible to control hierarchies of tasks using the full
dynamics of the robot. The main advantage is that it is possible to
express complicated behaviors directly at the task level with a strict
enforcement of hierarchies between tasks. It is, for example, useful
to ensure that a balancing task will take precedence over a task of
lower importance in case of conflicting goals. Early hierarchical
approaches are based on pseudo-inverse techniques~\cite{Sentis:2005}
and take inspiration directly from techniques used for
manipulators~\cite{Nakamura:1987}. However, pseudo-inverse-based
controllers are limited as they cannot properly handle inequality
constraints such as torque limits or friction cone constraints. More
recently, generalizations have been proposed
\cite{deLasa:2010hf,Saab:2013vr,Mansard:2012gy,Escande:2014en} that
naturally allow the inclusion of arbitrary types of tasks including
inequalities. The resulting optimization problems are phrased as
cascades of quadratic programs (QPs). Evaluation of their
applicability was done in simulation and it has been shown that these
algorithms are fast enough to be implemented in a real-time fast
control loop for inverse kinematics. It has also been argued that they
can be implemented fast enough for the use with inverse dynamics and
can work on robots with model-uncertainty, sensor noise and limited
torque bandwidth. But to the best of our knowledge, these controllers
have never been used as feedback-controllers on real torque-controlled
humanoids.

In \cite{Saab:2013vr}, the trajectories computed in simulation are
replayed on a real robot using joint space position control, but the
method is not used for feedback control in task-space on the robot.
This work is very interesting because it demonstrates that
trajectories generated by a hierarchical inverse dynamics are such
that they can be used on a real system. However, it is important to
note that this does not show that feedback control can be done using
these controllers. Indeed, when replaying trajectories, feedback is
reduced to joint level tracking. Therefore it is not possible to
directly control interaction forces during multi-contact tasks or to
close a feedback loop directly around the tasks of interests, for
example the center of gravity (CoG), that respects the desired
hierarchies.
It is worth mentioning that \cite{Hutter:2012uu} recently successfully
implemented a controller using the full dynamics of the robot and task
hierarchies on a torque controlled quadruped robot. The approach is
based on pseudo-inverses and not QPs which makes it potentially
inefficient to handle inequalities (e.g. friction cone constraints,
torque saturation, center of pressure constraints, etc...).

During balancing and walking tasks, an appropriate control of the CoG
is of major importance. Recently, it has been realized that the
control of both the linear (i.e. the CoG) and angular momentum of the
robot could be very beneficial for balancing and walking tasks. The
control of overall momentum was originally proposed in
~\cite{Kajita:2003gj} using a resolved rate control framework and it
was recently extended in~\cite{Lee:2012hb} where it was integrated
with an inverse dynamics controller. It has been shown in several
contributions \cite{Wensing:2013,Lee:2012hb} that the regulation of
momentum could be very powerful for control on humanoids. Despite the
growing popularity for momentum-based control approaches, there have
been very few evaluations of such techniques on a real humanoid
robot~\cite{Stephens:2010vu}. In~\cite{Stephens:2010vu} the
momentum-based control is computed using a simplification of the
optimization problem and does not necessarily generate the optimal
command. Moreover, the control command generated from inverse dynamics
is used in conjunction with a joint PD controller and not used as the
sole feedback controller of the system (i.e. there are two distinct
feedback pathways, one coming from the momentum control through
inverse dynamics and the other coming from desired joint positions at
the joint level). To the best of our knowledge a momentum-based
controller has never been evaluated either in a complete hierarchical
inverse dynamics framework or without additional joint PD
stabilization.

As advanced torque control techniques are developed there is a need to
evaluate these techniques on torque-controlled platform to assess
their capabilities and also their drawbacks. Such an evaluation is the
main goal of the paper. In a recent contribution~\cite{herzog:2014b},
we have demonstrated that hierarchical inverse dynamics controllers
could be efficiently used on a torque-controlled humanoid robot. In
particular, we demonstrated robust performance during balancing and
tracking tasks when using a momentum-based balance control approach.
We also proposed a method to simplify the optimization problem by
factoring the dynamics equations of the robot such that we could
significantly reduce computational time and achieve a \unit[1]{kHz}
control-loop.

\paragraph{Contribution}
In this contribution, we extend our preliminary work and present
extensive experimental evaluations. First, we show modifications we
applied to the algorithm, originally proposed
by~\cite{deLasa:2010hf,Kanoun:2011ey}, that were necessary to execute
it in a real-time feedback control setting for inverse dynamics tasks
(Section \ref{sec:hierarch_inv_dyn}).
%
We also propose a method to systematically compute the feedback gains
for the linear and angular momentum control task by using a linear
optimal control design approach (Section \ref{sec:lqr_momentum}).
This leads us to the main contribution of this paper, where we show
experiments with extensive quantitative analysis for various tasks
(Sections \ref{sec:experimental_setup} and \ref{sec:experiments}).
We show that the momentum-based controller with optimal feedback gains
can improve robot performance.
Balancing experiments in various conditions demonstrate performances
that are comparable to, if not better than, current state of the art
balancing algorithms, even when the robot is balancing on one foot.
Tracking and contact switching experiments also show the versatility
of the approach.
It is, to the best of our knowledge, the first demonstration of the
applicability of the methods proposed in \cite{deLasa:2010hf} or
\cite{Saab:2013vr} as feedback controllers on torque controlled
humanoids (i.e. without joint space PD control) with the use of a
momentum-based control approach. In the last section, we discuss the
experimental results as compared to the state of the art.

\section{Hierarchical Inverse Dynamics} \label{sec:hierarch_inv_dyn}

In this section, we detail our modeling assumptions, give a short
summary on how tasks can be formulated as desired closed-loop
behaviors and revisit the original solver
formulation~\cite{deLasa:2010hf}. In Section~\ref{sec:decomp} we then
propose a simplification to reduce the complexity of the original
formulation. The simplification is also applicable to any other
inverse dynamics formulation.

\subsection{Modelling Assumptions and Problem
  Formulation} \label{sec:task_formulation} In the following, we
describe the constraints and tasks that are considered by the
hierarchical inverse dynamics. They will all be written as affine
functions of joint and body accelerations, joint torques and contact
forces in order to formulate the control problem as a series of
quadratic programs. They constitute the variables that will be
optimized by the controller.

\paragraph{Rigid Body Dynamics}
Assuming rigid-body dynamics, we can write the equations of motion of
a robot as
\begin{equation}\label{eq:equations_of_motion}
  \mathbf{M}(\mathbf{q}) \ddot{\mathbf{q}} + \mathbf{N}(\mathbf{q},\dot{\mathbf{q}}) = \mathbf{S}^T \boldsymbol{\tau} + \mathbf{J}_c^T \boldsymbol{\lambda}
\end{equation}
where $\mathbf{q} =[\mathbf{q}_j^T~\mathbf{x}^T]^T$ denotes the
configuration of the robot. $\mathbf{q}_j \in \mathbb{R}^n$ is the
vector of joint positions and $\mathbf{x} \in \mathrm{SE}(3)$ denotes
the position and orientation of a frame fixed to the robot with
respect to an inertial frame (the floating base).
$\mathbf{M}(\mathbf{q})$ is the inertia matrix,
$\mathbf{N}(\mathbf{q},\dot{\mathbf{q}})$ is the vector of all
non-contact forces (Coriolis, centrifugal, gravity, friction, etc.),
$\mathbf{S} = [\mathbf{I}_{n\times n} \mathbf{0}]$ represents the
underactuation, $\boldsymbol{\tau}$ is the vector of commanded joint
torques, $\mathbf{J}_c$ is the Jacobian of the contact constraints and
$\boldsymbol{\lambda}$ are the generalized contact forces.

\paragraph{Contact constraints}
End effectors are constrained to remain stationary. We express the
constraint that the feet (or hands) in contact with the environment do
not move ($\mathbf{x}_c = const$) by differentiating it twice and
using the fact that
$\dot{\mathbf{x}}_c = \mathbf{J}_c \dot{\mathbf{q}}$. We get the
following equality constraint
\begin{equation}\label{eq:contact_constraints}
  \mathbf{J}_c \ddot{\mathbf{q}} + \dot{\mathbf{J}_c} \dot{\mathbf{q}} = \mathbf{0}.
\end{equation}

\paragraph{Center of pressure}
To ensure stationary contacts, the center of pressure (CoP) at each
end effector needs to reside in the interior of the end effector's
support polygon. This can be expressed as a linear inequality by
expressing the ground reaction force at the zero moment point.

\paragraph{Friction cone}
For the feet not to slip we constraint the ground reaction forces
(GRFs) to stay inside the friction cones. In our case, we approximate
the cones by pyramids to have linear inequality constraints in the
contact forces.

\paragraph{Torque and joint limits}
Especially important for generating control commands that are valid on
a robot is to take into account actuation limits
$\boldsymbol{\tau}_{min} \le \boldsymbol{\tau} \le
\boldsymbol{\tau}_{max}$.
The same is true for joint limits, which can be written as
$\ddot{\mathbf{q}}_{min} \le \ddot{\mathbf{q}}_j \le
\ddot{\mathbf{q}}_{max}$,
where the bounds are computed in the form
$\ddot{\mathbf{q}}_{min/max} \propto \text{tanh}(\mathbf{q -
  \ddot{\mathbf{q}}}_{min/max})$.

\paragraph{Motion and force control tasks}
Motion controllers can be phrased as
$\ddot{\mathbf{x}}_{ref} = \mathbf{J}_x \ddot{\mathbf{q}} +
\dot{\mathbf{J}}_x \dot{\mathbf{q}}$,
where $\mathbf{J}_x$ is the task Jacobian and
$\ddot{\mathbf{x}}_{ref}$ is a reference task acceleration that will
correspond to a desired closed-loop behavior (e.g. obtained from a
PD-controller). Desired contact forces can be directly expressed as
equalities on the generalized forces $\boldsymbol{\lambda}$. In
general, we assume that each control objective can be expressed as a
linear combination of $\ddot{\mathbf{q}}$, $\boldsymbol{\lambda}$ and
$\boldsymbol{\tau}$, which are the optimization variables of
our problem.\\

At every control cycle, the equations of motion
(Equation~\eqref{eq:equations_of_motion}), the constraints for
physical consistency (torque saturation, CoP constraints, etc.) and
our control objectives are all expressed as affine equations of the
variables
$\ddot{\mathbf{q}}, \boldsymbol{\lambda}, \boldsymbol{\tau}$. Tasks of
the same priority can then be stacked vertically into the form
\begin{alignat}{2}
  \label{eq:affine_ctrl_objs_eq}
  &\mathbf{A} \mathbf{y} + \mathbf{a}\leq 0, \\
  \label{eq:affine_ctrl_objs_ineq}
  &\mathbf{B} \mathbf{y}+ \mathbf{b} = 0,
\end{alignat}
where
$\mathbf{y} =
[\ddot{\mathbf{q}}^T~\boldsymbol{\lambda}^T~\boldsymbol{\tau}^T]^T$,
$\mathbf{A} \in \mathbb{R}^{m \times (2n+6+6c)}$,
$\mathbf{a} \in \mathbb{R}^{m}$,
$\mathbf{B} \in \mathbb{R}^{k \times (2n+6+6c)}$,
$\mathbf{b} \in \mathbb{R}^{k}$ and $m, k \in \mathbb{N}$ the overall
task dimensions and $n \in \mathbb{N}$ the number of robot DoFs.
$c \in \mathbb{N}$ is the number of constrained end effectors.

The goal of the controller is to find $\ddot{\mathbf{q}}$,
$\boldsymbol{\lambda}$ and $\boldsymbol{\tau}$ (and therefore a
control command) that satisfies these objectives as well as possible.
Objectives will be stacked into different priorities, with the highest
priority in the hierarchy given to physical consistency. In a lower
priority, we will express balancing and motion tracking tasks and we
will put tasks for redundancy resolution in the lowest priorities.

\subsection{Hierarchical Tasks \& Constraints
  Solver}\label{sec:hierarchical_inv_dyn_solver}
The control objectives and constraints in Equations
\eqref{eq:affine_ctrl_objs_eq} and \eqref{eq:affine_ctrl_objs_ineq}
might not have a common solution, but need to be traded off against
each other. In case of a push, for instance, the objective to
decelerate the CoG might conflict with a swing foot task. A tradeoff
can be expressed in form of slacks on the expressions in
Equations~\eqref{eq:affine_ctrl_objs_eq},\eqref{eq:affine_ctrl_objs_ineq}.
The slacks are then minimized in a quadratic program. We propose an
algorithm that is a combination of the methods originally proposed
in~\cite{deLasa:2010hf,Kanoun:2011ey}.

\begin{alignat}{2}
  \label{eq:single_qp}
  & \underset{\mathbf{y}, \mathbf{v}, \mathbf{w}}{\text{min.}}~
  & 		 &\|\mathbf{v}\|^2 + \|\mathbf{w}\|^2 + \epsilon \|\mathbf{y}\| \\
  \label{eq:single_qp_ineq}
  & \text{s.t.} & &~ \mathbf{V}(\mathbf{A} \mathbf{y} + \mathbf{a})\leq \mathbf{v}, \\
  \label{eq:single_qp_eq}
  & & & \mathbf{W}(\mathbf{B} \mathbf{y}+ \mathbf{b}) = \mathbf{w},
\end{alignat}
where matrices
$\mathbf{V} \in \mathbb{R}^{m \times m}, \mathbf{W} \in \mathbb{R}^{k
  \times k}$
weigh the cost of constraints against each other and
$\mathbf{v} \in \mathbb{R}^m, \mathbf{w} \in \mathbb{R}^k$ are slack
variables. Note that $\mathbf{v}, \mathbf{w}$ are not predefined, but
part of the optimization variables. The objective is regularized by a
small value $\epsilon$ (typically $10^{-4}$), which ensures positive
definiteness of the objective hessian. In the remainder, we write the
weighted tasks using
$\bar{\mathbf{A}}=\mathbf{V}\mathbf{A},
\bar{\mathbf{a}}=\mathbf{V}\mathbf{a},
\bar{\mathbf{B}}=\mathbf{W}\mathbf{B},
\bar{\mathbf{b}}=\mathbf{W}\mathbf{b}$.

Although $\mathbf{W}, \mathbf{V}$ allow us to trade-off control
objectives against each other, strict prioritization cannot be
guaranteed with the formulation in Equation~\eqref{eq:single_qp}. For
instance, we might want to trade off tracking performance of tasks
against each other, but we do not want to sacrifice physical
consistency of a solution at any cost. In order to guarantee
prioritization, we solve a sequence of QPs, in which a QP with
constraints imposed by lower priority tasks is optimized over the set
of optimal solutions of higher priority tasks as proposed
by~\cite{Kanoun:2011ey}. Given one solution
$(\mathbf{y}_r^*, \mathbf{v}_r^*)$ for the QP of priority $r$, all
remaining optimal solutions $\mathbf{y}$ in that QP are expressed by
the equations

\begin{eqnarray}
  \label{eq:optimals_constr_1}
  &\mathbf{y} = \mathbf{y}_r^* + \mathbf{Z}_r\mathbf{u}_{r+1}, \\
  \label{eq:optimals_constr_2}
  &\bar{\mathbf{A}}_r \mathbf{y}+ \bar{\mathbf{a}}_r  \leq \mathbf{v}_r^*, \\ \nonumber
  &\dots \\
  &\bar{\mathbf{A}}_1 \mathbf{y}+ \bar{\mathbf{a}}_1 \leq  \mathbf{v}_1^*, \nonumber
\end{eqnarray} 
where $\mathbf{Z}_r \in \mathbb{R}^{(2n+6+6c) \times z_r}$ represents
a surjective mapping into the nullspace of all previous equalities
$\bar{\mathbf{B}}_r, \dots$, $\bar{\mathbf{B}}_{1}$ and
$\mathbf{u}_r \in \mathbb{R}^{z_r}$ is a variable that parameterizes
that nullspace. We compute $\mathbf{Z}_r$ from a Singular Value
Decomposition (SVD).
%
%
With this nullspace mapping we reduce the number of variables from one
hierarchy level to the next by the number of locked degrees of
freedom. In our implementation the SVD is computed in parallel with
the QP at priority level $r-1$ and rarely finishes after the QP, i.e.
it adds only a negligible overhead.

Now, we can express a QP of the next lower priority level $r+1$ and
additionally impose the constraints in
Equations~\eqref{eq:optimals_constr_1},~\eqref{eq:optimals_constr_2}
in order to optimize over $\mathbf{y}$ without violating optimality of
higher priority QPs:

\begin{alignat}{2}
  \label{eq:hierarch_qp}
  & \underset{\mathbf{u}_{r+1}, \mathbf{v}_{r+1}}{\text{min.}}~ &
  &\|\bar{\mathbf{B}}_{r+1} (\mathbf{y}_r^* +
  \mathbf{Z}_r\mathbf{u}_{r+1}) + \bar{\mathbf{b}}_{r+1}\| + \\
  \nonumber
  & 		 &&\|\mathbf{v}_{r+1}\| + \epsilon \|\mathbf{y}\|\\
  & ~~~\text{s.t.}
  &  &\bar{\mathbf{A}}_{r+1} ( \mathbf{y}_r^* + \mathbf{Z}_r\mathbf{u}_{r+1},) +\bar{\mathbf{a}}_{r+1} \leq \mathbf{v}_{r+1}, \nonumber \\
  \label{eq:hierarch_qp_previneq}
  &&&~~~~~~\bar{\mathbf{A}}_r ( \mathbf{y}_r^* + \mathbf{Z}_r\mathbf{u}_{r+1},)+ \bar{\mathbf{a}}_r \leq \mathbf{v}_r^*,  \\
  &&&~~~~~~\hspace{2cm} \dots \nonumber \\
  &&&~~~~~~\bar{\mathbf{A}}_1 ( \mathbf{y}_r^* +
  \mathbf{Z}_r\mathbf{u}_{r+1},)+ \bar{\mathbf{a}}_1 \leq
  \mathbf{v}_1^*, \nonumber
\end{alignat}
where we wrote the QP as in Equation~\eqref{eq:single_qp} and
substituted $\mathbf{w}$ into the objective function. In order to
ensure that we optimize over the optimal solutions of higher priority
tasks, we added Equation~\eqref{eq:optimals_constr_2} as an additional
constraint and substituted Equation~\eqref{eq:optimals_constr_1} into
Equations~\eqref{eq:hierarch_qp}-\eqref{eq:hierarch_qp_previneq}. This
allows us to solve a stack of hierarchical tasks recursively as
originally proposed by~\cite{Kanoun:2011ey}.
%
%
Right-multiplying $\mathbf{Z}_r$ to inequality matrices
$\bar{\mathbf{A}}_{r}$ creates zero rows for constraints that do not
have degrees of freedom left. For example, after the GRFs are decided,
CoP and friction constraints become obsolete. This way the number of
inequalities reduces potentially from one QP to the other. Note that
this optimization algorithm is guaranteed to find the optimal solution
in a least-squares sense while satisfying priorities.

With this formulation we combine the two benefits of having
inequalities in all hierarchical levels~\cite{Kanoun:2011ey} and
reducing the number of variables from one QP to the
other~\cite{deLasa:2010hf}.

\subsection{Decomposition of Equations of Motion}\label{sec:decomp}

Hierarchical inverse dynamics approaches usually have in common that
consistency of the variables with physics, i.e. the equations of
motion, need to be ensured. In~\cite{deLasa:2010hf} these constraints
are expressed as equality constraints (with slacks) resulting in an
optimization problem over all variables
$\ddot{\mathbf{q}}, \boldsymbol{\tau}, \boldsymbol{\lambda}$.
In~\cite{Mansard:2012gy} a mapping into the nullspace of
Equation~\eqref{eq:equations_of_motion} is obtained from a SVD on
Equation~\eqref{eq:equations_of_motion}. In both cases, complexity can
be reduced as we will show in the following. We decompose the
equations of motion as

\begin{eqnarray}\label{eq:decomp_eq_motion_upper}
  \mathbf{M_u(\mathbf{q})} \ddot{\mathbf{q}} +
  \mathbf{N_u}(\mathbf{q},\dot{\mathbf{q}}) 
  &=& 
      \boldsymbol{\tau} +
      \mathbf{J}_{c,u}^T 
      \boldsymbol{\lambda}, \\
  \label{eq:decomp_eq_motion_lower}
  \mathbf{M_l(\mathbf{q})} \ddot{\mathbf{q}}  +
  \mathbf{N_l}(\mathbf{q},\dot{\mathbf{q}})
  &=&
      \mathbf{J}_{c,l}^T
      \boldsymbol{\lambda}
\end{eqnarray}
where Equation~\eqref{eq:decomp_eq_motion_upper} is just the first $n$
equations of Eq. \eqref{eq:equations_of_motion} and
Equation~\eqref{eq:decomp_eq_motion_lower} is the last $6$ equations
related to the floating base. The latter equation can then be
interpreted as the Newton-Euler equations of the whole system
\cite{Wieber:2006up}. They express the change of momentum of the robot
as a function of external forces. A remarkable feature of the
decomposition in
Equations~\eqref{eq:decomp_eq_motion_upper},~\eqref{eq:decomp_eq_motion_lower}
is that the torques $\boldsymbol{\tau}$ only occur in
Equation~\eqref{eq:decomp_eq_motion_upper} and are exactly determined
by $\ddot{\mathbf{q}}, \boldsymbol{\lambda}$ in the form

\begin{equation} \label{eq:torque_substitution} \boldsymbol{\tau} =
  \mathbf{M_u(\mathbf{q})}\ddot{\mathbf{q}} +
  \mathbf{N_u}(\mathbf{q},\dot{\mathbf{q}}) - \mathbf{J}_{c,u}^T
  \boldsymbol{\lambda}
\end{equation}

Since $\boldsymbol{\tau}$ is linearly dependent on
$\ddot{\mathbf{q}}, \boldsymbol{\lambda}$, for any combination of
accelerations and contact forces there always exists a solution for
$\boldsymbol{\tau}$. It is given by
Equation~\eqref{eq:torque_substitution}. Therefore, it is only
necessary to use Equation~\eqref{eq:decomp_eq_motion_lower} as a
constraint for the equations of motion during the optimization (i.e.
the evolution of momentum is the only constraint).

Because of the linear dependence, all occurrences of
$\boldsymbol{\tau}$ in the problem formulation (i.e. in
Equations~\eqref{eq:affine_ctrl_objs_eq}-\eqref{eq:affine_ctrl_objs_ineq})
can be replaced with Equation~\eqref{eq:torque_substitution}. This
reduces the number of variables in the optimization from $(2n+6+6c)$
to $(n+6+6c)$. This decomposition thus eliminates as many variables as
there are DoFs on the robot. This simplification is crucial to reduce
the time taken by the optimizer and allowed us to implement the
controller in a \unit[1]{kHz} feedback control loop.
\begin{remark}
  The simplification that we propose\footnote{We originally proposed
    the simplification in a technical note~\cite{herzog:2013a}.} can
  appear trivial at first sight. However, it is worth mentioning that
  such a decomposition is always ignored in related work despite the
  need for computationally fast
  algorithms~\cite{deLasa:2010hf},~\cite{Mansard:2012gy},~\cite{Stephens:2010vu}.
\end{remark}

\subsection{Solution to the first priority}
Since we are interested in writing inverse dynamics controllers, we
set the highest priority tasks to always be the Newton-Euler Equations
(Equation ~\eqref{eq:decomp_eq_motion_lower}) together with torque
saturation constraints. We then need to find the space of solutions
for equations
\begin{eqnarray}
  &\mathbf{B}_1\mathbf{y} + \mathbf{b}_1 = 0\\
  &-\boldsymbol{\tau}_{sat} \leq \boldsymbol{\tau}(\mathbf{y}) \leq \boldsymbol{\tau}_{sat}
\end{eqnarray}
with $\boldsymbol{\tau}(\mathbf{y})$ given by
Equation~\eqref{eq:torque_substitution},
$\mathbf{B}_1 = \begin{bmatrix} \mathbf{M_l} ~
  -\mathbf{J}_{c,l}^T \end{bmatrix}$
and $\mathbf{b}_1 = \mathbf{N_l}$. In this case, we can obtain the
space of solutions (cf. Equation~\eqref{eq:optimals_constr_1}) without
having to solve a QP. A trivial solution can be readily obtained, thus
reducing computation time.
Indeed, it is always possible to satisfy the equations of motion
together with the torque saturation constraints exactly by choosing
$\boldsymbol{\tau} = \boldsymbol{\lambda} = \mathbf{0}$ and resolving
for $\ddot{\mathbf{q}}=-\mathbf{M^{-1}}\mathbf{N}$ using
Equation~\eqref{eq:equations_of_motion}. The resulting solution will
be in the set of minimizers, i.e.

\begin{eqnarray}\label{eq:first_qp_solution_existance}
  \exists \mathbf{u}_1: &&\mathbf{y} = \begin{bmatrix}\mathbf{\ddot{q}}  \\ \mathbf{0}\end{bmatrix} = 
  -\mathbf{B^\dag}_1\mathbf{b}_1 + \mathbf{Z}_1\mathbf{u}_1,\\
  \label{eq:first_qp_torque_zero}
                        &\wedge& \boldsymbol{\tau}(\mathbf{y}) = \mathbf{0}
\end{eqnarray} 

with $\mathbf{Z}_1$ computed as described in
Section~\ref{sec:hierarchical_inv_dyn_solver}. We can then obtain
$ \mathbf{y}_1^*= -\mathbf{B^\dag}_1\mathbf{b}_1, \mathbf{v}_1^*= 0$,
which is required to construct the QP for priority $r=2$. Although
$\mathbf{y}_1^*$ may violate torque saturation constraints,
Equations~\eqref{eq:first_qp_solution_existance},~\eqref{eq:first_qp_torque_zero}
guarantee that an admissible $\mathbf{y}$ can always be found and will
be found in the following QPs. With this choice of $\mathbf{y}_1^*$
there is no need to invert $\mathbf{M}$. Note that $\mathbf{B}_1$
represents the Newton-Euler equations of the system and is always of
full row rank \footnote{The part of $\mathbf{M_l}$ multiplying the
  base acceleration is always full rank.} and thus computing
$\mathbf{B^\dag}_1$ requires only inverting the $6\times 6$ sized
positive definite matrix $\mathbf{B}_1\mathbf{B}_1^T$.
By designing the first hierarchy level in this way, we can improve
computation time by avoiding to solve the first QP while already
reducing the size of the problem by 6 variables for the next priority.

\section{Linear and angular momentum
  regulation}\label{sec:lqr_momentum}
As we mentioned in the introduction, we are interested in writing
desired feedback behaviors using hierarchical inverse dynamics and
more specifically, we are interested in controlling the linear and
angular momentum of the robot. The feedback controller that regulates
momentum is often written as a PID controller with hand-tuned gains.
Such control design does not take into account the coupling between
linear and angular momentum during a multi-contact task and can
potentially lead to a controller which is sub-optimal and difficult to
tune.

In this section, we write the momentum regulation problem as a force
control task and then use a simple LQR design to compute a linear
optimal feedback control law. This feedback law is then used to
compute a desired closed-loop behavior in the hierarchical inverse
dynamics controller. The advantage of such design is that it fully
exploits multi-contacts and momentum coupling while significantly
simplifying the design of the controller by reducing the number of
open parameters.

\subsection{Linear and angular momentum models}
The control of momentum and CoG is inherently both a kinematic and a
force task. Indeed, using the centroidal momentum matrix
\cite{Orin:2008ge}, one can find a linear mapping between the overall
robot momentum and the robot joint and pose velocities

\begin{equation}
  \label{eq:momentum_mat_eq}
  \mathbf{h} = \mathbf{H_G}(\mathbf{q})\dot{\mathbf{q}} 
\end{equation}
where $\mathbf{h} = [\mathbf{h}_{lin}^T ~\mathbf{h}_{ang}^T]^T $ is
the system linear and angular momentum expressed at the CoG. The
matrix $\mathbf{H_G}$ is called the centroidal momentum matrix. The
derivative of Equation~\eqref{eq:momentum_mat_eq} allows us to express
the rate of change of the momentum and the CoG
\begin{eqnarray}
  \dot{\mathbf{x}}_{cog} &=& \frac{1}{m}\mathbf{h}_{lin} \nonumber\\
  \dot{\mathbf{h}} &=& \mathbf{H_G}\ddot{\mathbf{q}} + \dot{\mathbf{H}}_\mathbf{G}\dot{\mathbf{q}} \label{eq:mom_rate_ddq}
\end{eqnarray}
This formulation has been often used in a resolved acceleration scheme
where the centroidal momentum matrix is viewed as the task Jacobian
(e.g. in \cite{Lee:2012hb}).

Using the Newton-Euler equations, the total change of momentum can
also be written in terms of the external forces
\begin{eqnarray}
  \dot{\mathbf{x}}_{cog} &=& \frac{1}{m}\mathbf{h}_{lin} \nonumber \\
  \dot{\mathbf{h}}  &=& \begin{bmatrix}
    \mathbf{I}_{3\times 3} & \mathbf{0}_{3 \times 3} &\ldots	
    \\
    \label{eq:mom_rate_frcs} [\mathbf{x}_i -
    \mathbf{x}_{cog}]_{\times} &\mathbf{I}_{3\times 3}& \ldots
  \end{bmatrix}
                                                        \boldsymbol{\lambda} + 
                                                        \begin{bmatrix}
                                                          m\mathbf{g}
                                                          \\
                                                          \mathbf{0}
                                                        \end{bmatrix},
\end{eqnarray}
where $m\mathbf{g}$ is the gravitational force, $\boldsymbol{\lambda}$
the vector of generalized external forces, $[\centerdot]_{\times}$
maps a vector to a skew symmetric matrix, s.t.
$[\mathbf{x}]_{\times}\boldsymbol{\lambda} = \mathbf{x} \times
\boldsymbol{\lambda}$
and $\mathbf{x}_i$ is the position of the $i^{th}$ contact point.

We see that the rate of momentum change can equivalently be written
either as a kinematic task (i.e. a function of $\mathbf{\ddot{q}}$ as
in Equation~\eqref{eq:mom_rate_ddq}) or a force task (i.e. a function
of $\boldsymbol{\lambda}$ as in Equation~\eqref{eq:mom_rate_frcs}).
The matrix in front of $\mathbf{\ddot{q}}$ or $\boldsymbol{\lambda}$
is viewed as the Jacobian of the task.

In general, deriving a momentum control law with
Equation~\eqref{eq:mom_rate_frcs} might be better because we do not
have to compute $\dot{\mathbf{H}}_G$, which usually is acquired
through numerical derivation and might suffer from magnified noise. In
addition, in Equation~\eqref{eq:mom_rate_frcs} external forces can be
interpreted as the control inputs of the system, which is a useful
interpretation for control design, as we explain below.

\subsection{LQR design for momentum control}
A desired momentum behavior is typically achieved using a PD control
law, for example
\begin{eqnarray*}
  \dot{\mathbf{h}}_{des} = \mathbf{P}
  \begin{bmatrix}
    m (\mathbf{x}_{ref} - \mathbf{x}_{cog}) \\ \mathbf{0}
  \end{bmatrix} + && \mathbf{D} (\mathbf{h}_{ref} - \mathbf{h}) +
  \dot{\mathbf{h}}_{ref}
\end{eqnarray*}
where $\mathbf{h}_{ref}$ and $\mathbf{x}_{ref}$ are reference momentum
and CoG trajectories. Using Equation~\eqref{eq:mom_rate_ddq}, a
desired closed-loop behavior is then added in the hierarchical inverse
dynamics as
\begin{eqnarray}
  \mathbf{H_G}\ddot{\mathbf{q}} + && \dot{\mathbf{H}}_\mathbf{G}\dot{\mathbf{q}}  \\
                                  &&=\mathbf{P}
                                     \begin{bmatrix}
                                       m (\mathbf{x}_{ref} -
                                       \mathbf{x}_{cog}) \\ \mathbf{0}
                                     \end{bmatrix} + \nonumber
                                     \mathbf{D} (\mathbf{h}_{ref} -
                                     \mathbf{h}) +
                                     \dot{\mathbf{h}}_{ref}
  \label{eq:momentum_diag_gain_task}
\end{eqnarray}

There are, however, several issues with such an approach. First, the
tuning of the PD controller can be problematic. In our experience, on
the real robot it is necessary to have different gains for different
contact configurations to ensure proper tracking which leads to a time
consuming process with many open parameters. Second, such a controller
does not exploit the coupling between linear and angular momentum rate
of change that is expressed in Equation~\eqref{eq:mom_rate_frcs}.

We propose to use the model of Equation~\eqref{eq:mom_rate_frcs} to
compute optimal feedback gains. We linearize the dynamics and compute
a LQR controller by selecting a desired performance cost.
We find a control law of the form
\begin{eqnarray}
  \boldsymbol{\lambda} = -\mathbf{K}
  \left[ \begin{array}{c}
           \mathbf{x}_{cog} \\
           \mathbf{h}
         \end{array} \right] + \mathbf{k}(\mathbf{x}_{ref},\mathbf{h}_{ref}) \label{eq:lambda_momentum}
\end{eqnarray}
that contains both feedback and feedforward terms. A desired
closed-loop behavior for the momentum that appropriately takes into
account the momentum coupling is then computed. The desired task used
in the hierarchical inverse dynamics controller is then written as
\begin{eqnarray}
  \begin{bmatrix}
    \mathbf{I}_{3\times 3} & \mathbf{0}_{3 \times 3} &\ldots
    \\
    [\mathbf{x}_i - \mathbf{x}_{cog}]_{\times} &\mathbf{I}_{3\times
      3}& \ldots
  \end{bmatrix}
          \left (
          \boldsymbol{\lambda} + 
          \mathbf{K}
          \left[ \begin{array}{c}
                   \mathbf{x}_{cog} \\
                   \mathbf{h}
                 \end{array} \right]
  -  \mathbf{k}
  \right) = 0
  \label{eq:lqr_task}
\end{eqnarray}

We project the control $\boldsymbol{\lambda}$ into the momentum space
such that we can use the available redundancy during multi-contact
tasks to optimize the internal forces further. It would not be
possible if we used directly Equation~\eqref{eq:lambda_momentum}.

The proposed approach takes into account the coupling between linear
and angular momentum, which will prove beneficial in the experimental
section. Moreover, we specify the performance cost once and for all
and the feedback gains are computed optimally for every contact and
pose configuration of the robot at a low computational cost. In our
experience, it drastically simplified the application on the real
robot.

\begin{remark}
  In our experiments, we use an infinite horizon LQR design and
  compute gains for key poses of the robot, one for each contact
  configurations. During a contact transition we interpolate between
  the old and new set of gains to ensure continuous control commands.
  This solution is not ideal from a theoretical point of view as the
  interpolation does not guarantee stable behavior, but it works well
  in practice. Indeed, the contact transitions are very fast and all
  the trajectories were planned in advance. It would also be
  straightforward to linearize the dynamics at every control sequence
  and use a receding horizon controller with time-varying gains to
  allow online replanning of desired trajectories.
%
\end{remark}

\section{Experimental Setup}\label{sec:experimental_setup}
In this section, we detail the experimental setup, the low-level
feedback torque control, the state estimation algorithm and the
limitations of the hardware. These details are important in order to
understand the stren\-gths and limitations of the presented
experiments. They should also ease the reproduction of the
experimental results on other platforms.

\subsection{Sarcos Humanoid Robot}

\begin{figure}
  \centering
  \includegraphics[width=0.5\linewidth]{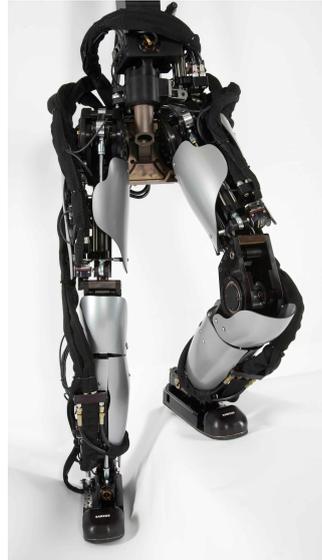}
  \caption{The lower part of the Sarcos Humanoid. (Credit:Luke Fisher Photography)}\label{fig:biped}
\end{figure}

The experiments were done on the lower part of the Sarcos Humanoid
Robot \cite{Cheng:2008tw}, shown in Figure~\ref{fig:biped}. It
consists of two legs and a torso. The legs have 7 DoFs each and the
torso has 3 DoFs. Given that the torso supports a negligible mass,
because it is not connected to the upper body of the robot and its
motion does not significantly influence the dynamics, we froze these
DoFs during the experiments. The legs of the robot are 0.82m high.
Each foot is 0.09m wide and 0.25m long. Note also that the front of
the foot is made of a passive joint that is rather flexible, located
10cm before the tip of the foot. Moving the CoP across this link makes
the foot bend and causes the robot to fall. This makes the effectively
used part of the sole rather small for a biped. The total robot mass
is 51kg.

The robot is actuated with hydraulics and each joint consists of a
Moog Series 30 flow control servo valve that moves a piston. Attached
to the piston is a load cell to measure the force at the piston. A
position sensor is also located at each joint. Each foot has a 6-axis
force sensor and we mounted an IMU on the pelvis of the robot from
which we measure angular velocities and linear accelerations of the
robot in an inertial frame. An offboard computer sends control
commands to the robot and receives sensor information in real-time at
\unit[1]{kHz}. The control commands consist of the desired current
applied to each valve. We used a computer running a linux kernel
patched with Xenomai 2.6.3 for real-time capabilities.

\subsection{Low-level torque control}
For each actuator, we implemented a torque feedback controller that
ensures that each joint produces the desired force generated by the
hierarchical inverse dynamics controller. The controller essentially
computes desired flow directly in terms of valve current. The
controller we implemented is very much inspired from the work in
\cite{Boaventura:2012tc,Boaventura:2012va}, with the difference that
we implemented a simpler version where piston velocity feedback has a
constant gain. The constant gain allows us to avoid the computation of
the piston chamber sizes and the measurement of the pressure inside.
The control law is
\begin{equation}
  v = PID(F_{des},F) + K \dot{x}_{piston} + d
\end{equation}
where $v$ is the valve command, $PID$ is a PID controller according to
desired force command and force measured from the load cells, $K$ is a
positive gain, $\dot{x}_{piston}$ is the piston velocity (computed
from the joint velocity and the kinematic model) and $d$ is a constant
bias.

\begin{figure}
  \centering
  \includegraphics[width=\linewidth]{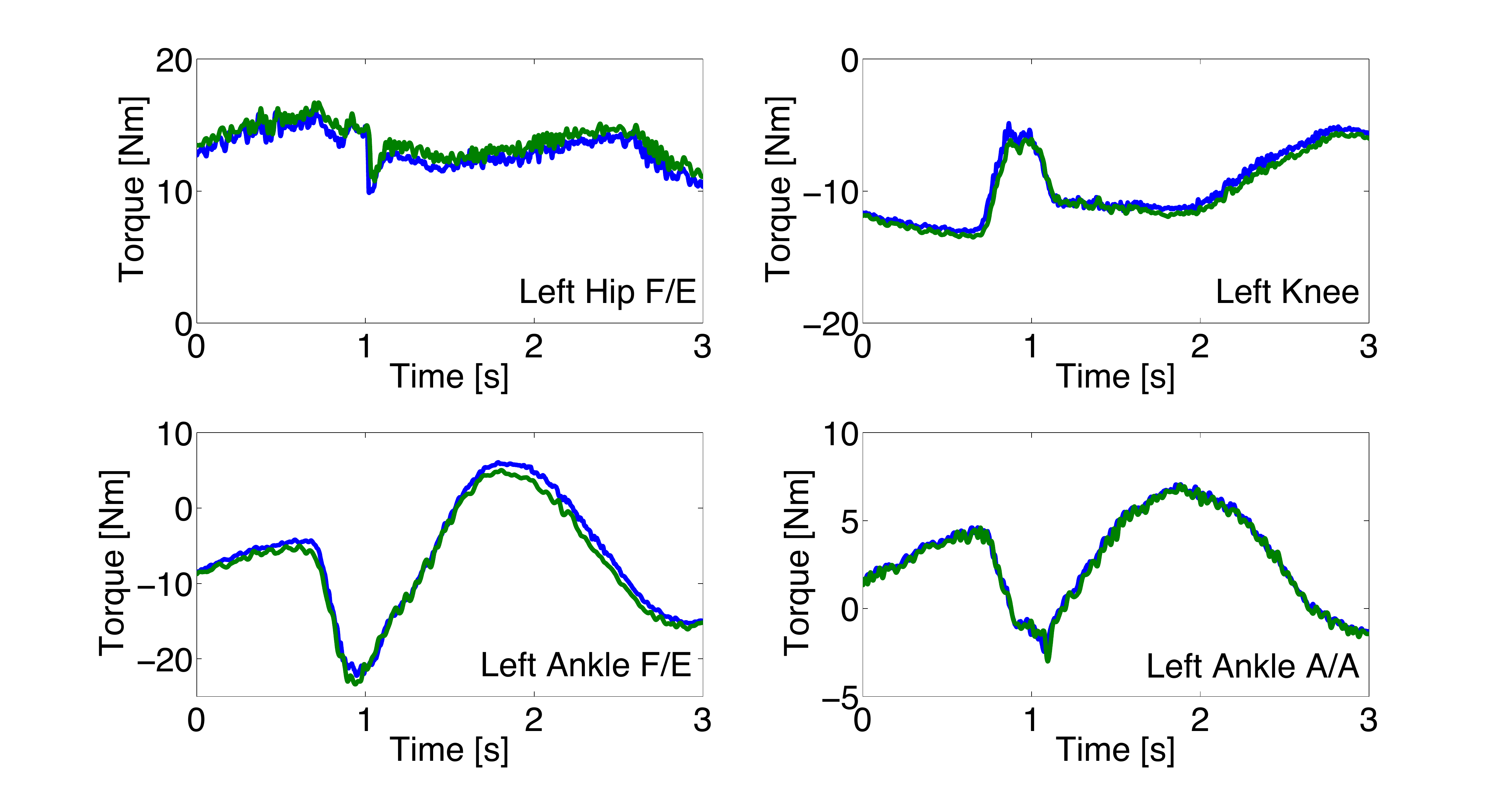}
  \caption{Example of torque tracking performance during a balancing
    experiment. The left hip flexion/extension, left knee and left
    ankle flexion/extension and adduction/abduction joints are shown.
    Both desired (blue) and actual (green) torques are
    shown.}\label{fig:torque_tracking}
\end{figure}

This controller design allowed us to achieve good torque tracking
performance. It is important to note that such performance was
necessary to achieve good performance in the hierarchical inverse
dynamics controller. Figure~\ref{fig:torque_tracking} illustrates the
torque tracking performance during a balancing experiment.

\subsection{State estimation}
An accurate estimation of the floating base pose and twist is
important for a good performance of the inverse dynamics controller.
We used a recently developed approach \cite{Rotella:2014tg} based from
the ideas in \cite{Bloesch:2012wu}. The estimation uses an extended
Kalman filter that fuses information from both the IMU and leg
kinematics. The filter handles contact switching and makes no
assumption about the gait or contact location in the world but only
uses the knowledge that a leg is in contact. It also has favorable
observability characteristics which make it particularly convenient
for our experiments. More details on the filter can be found in
\cite{Rotella:2014tg}.

\subsection{Dynamic model}
Our dynamic model is based on the CAD model of the robot. This means
that it is not very accurate as it does not take into account the
contribution of the hydraulic hoses, the electronics or any type of
friction in the model. We expect to have even better performance once
we perform a good identification of the dynamics
\cite{Ayusawa:2014,Mistry:2009dh} but it is interesting to note that
good results with hierarchical inverse dynamics can be obtained
without a perfect dynamic model. It demonstrates that these methods
are robust to model uncertainty in a compilation of balancing and
tracking tasks.

\subsection{Experimental tools}
\begin{figure}
  \centering
  \includegraphics[width=\linewidth]{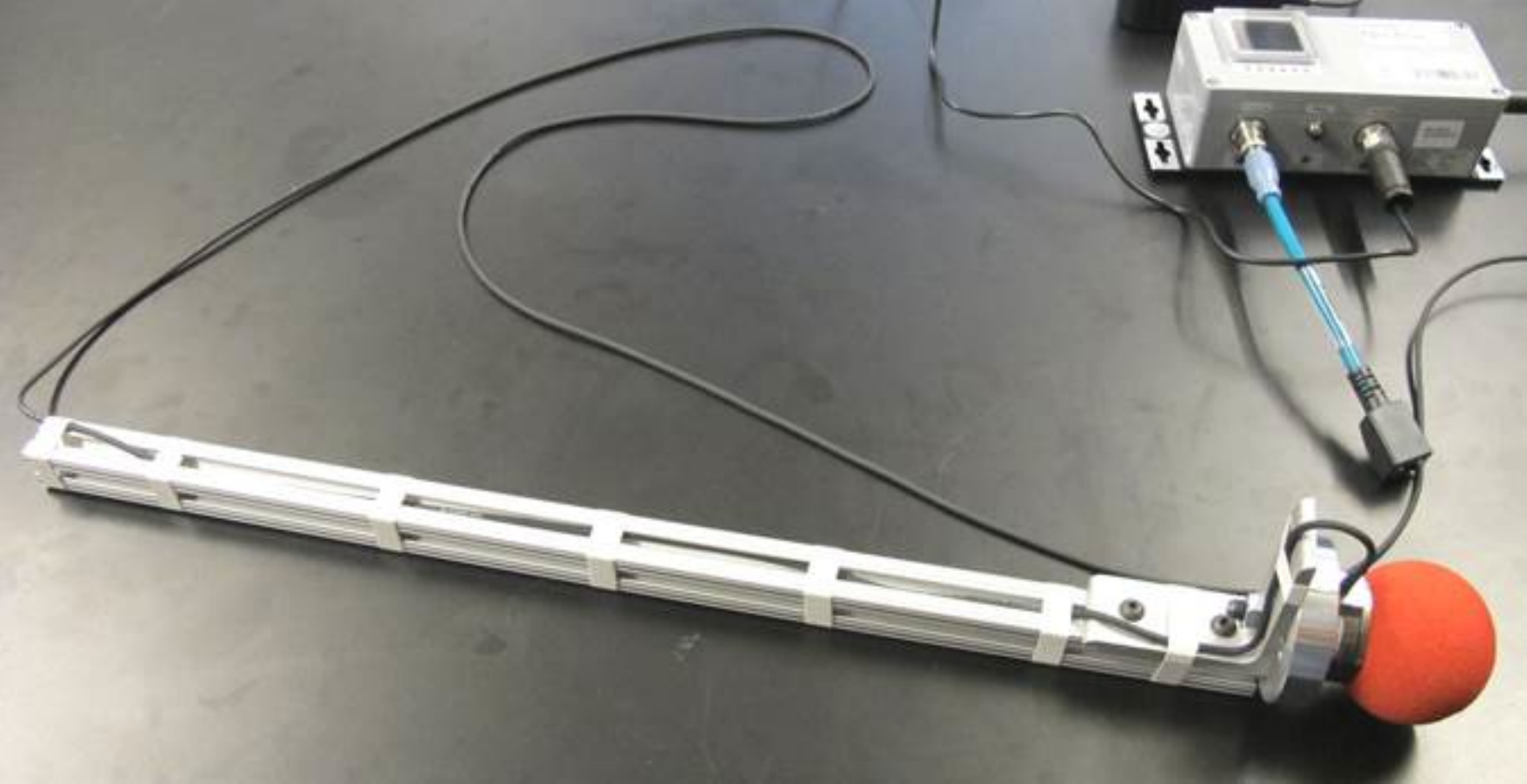}
  \includegraphics[width=\linewidth]{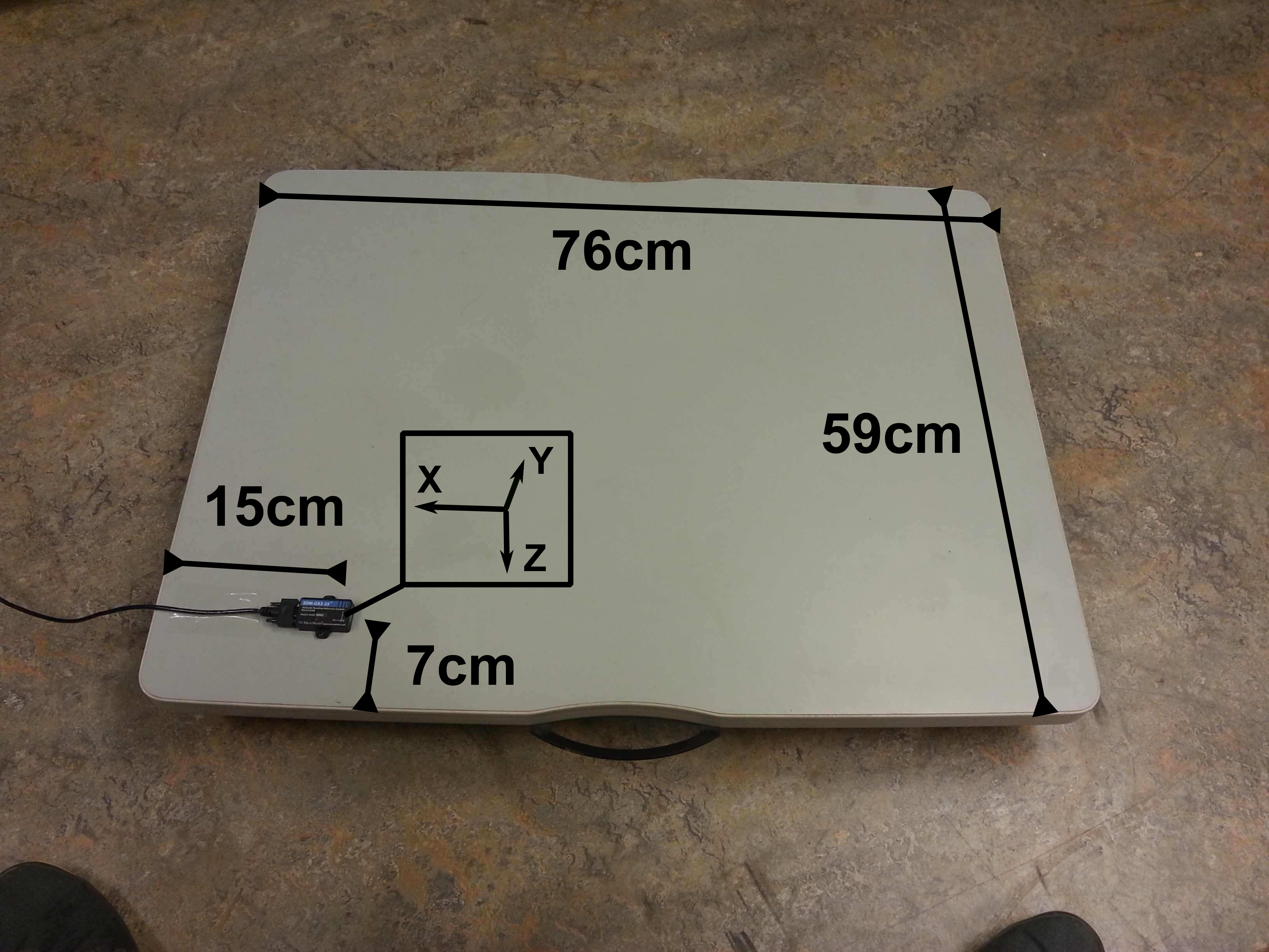}
  \includegraphics[width=\linewidth]{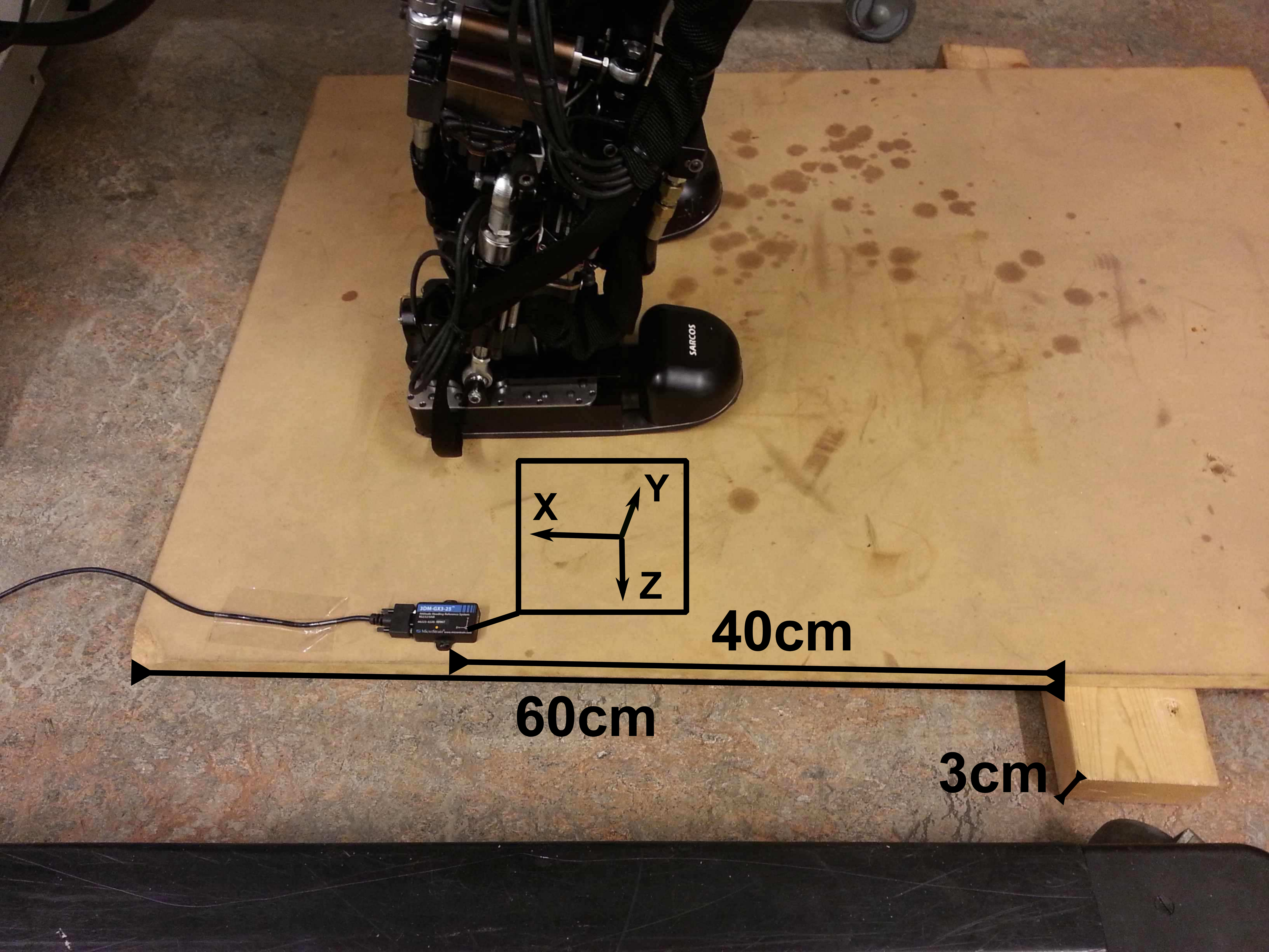}
  \caption{We attach a FTN-Mini 45 force sensor to a stick (top) to
    measure forces during pushes. In some of our experiments the robot
    stands on a rolling platform (middle) or a balancing board
    (bottom). In both scenarios an IMU is attached to the platform,
    which allows measuring linear accelerations and angular
    velocities, when disturbances are applied. The black box shows the
    internal frame of the IMU.}\label{fig:experiment_utils}
\end{figure}

For our experiments we use different tools to generate and measure
disturbances on the robot. We built a push stick that has a FTN-Mini
45 force sensor attached. It measures the applied force over time when
we push the robot. We conduct experiments where the robot is standing
on a rolling platform or a tilting platform that is put on top of a
beam. In both scenarios we attached a Microstrain 3DM-GX3-25 IMU (See
Figure~\ref{fig:experiment_utils}) to the plate that the robot is
standing on in order to measure linear acceleration and angular
velocities of the platform when a disturbance is applied. These
sensors are connected to the controlling computer together with the
robot sensors, which allows for easy synchronization of the readings.
We use real-time ethernet for the force sensor and real-time USB for
the IMU.

\section{Experiments}\label{sec:experiments}
\begin{figure*}[t]
  \centering
  \includegraphics[width=\linewidth]{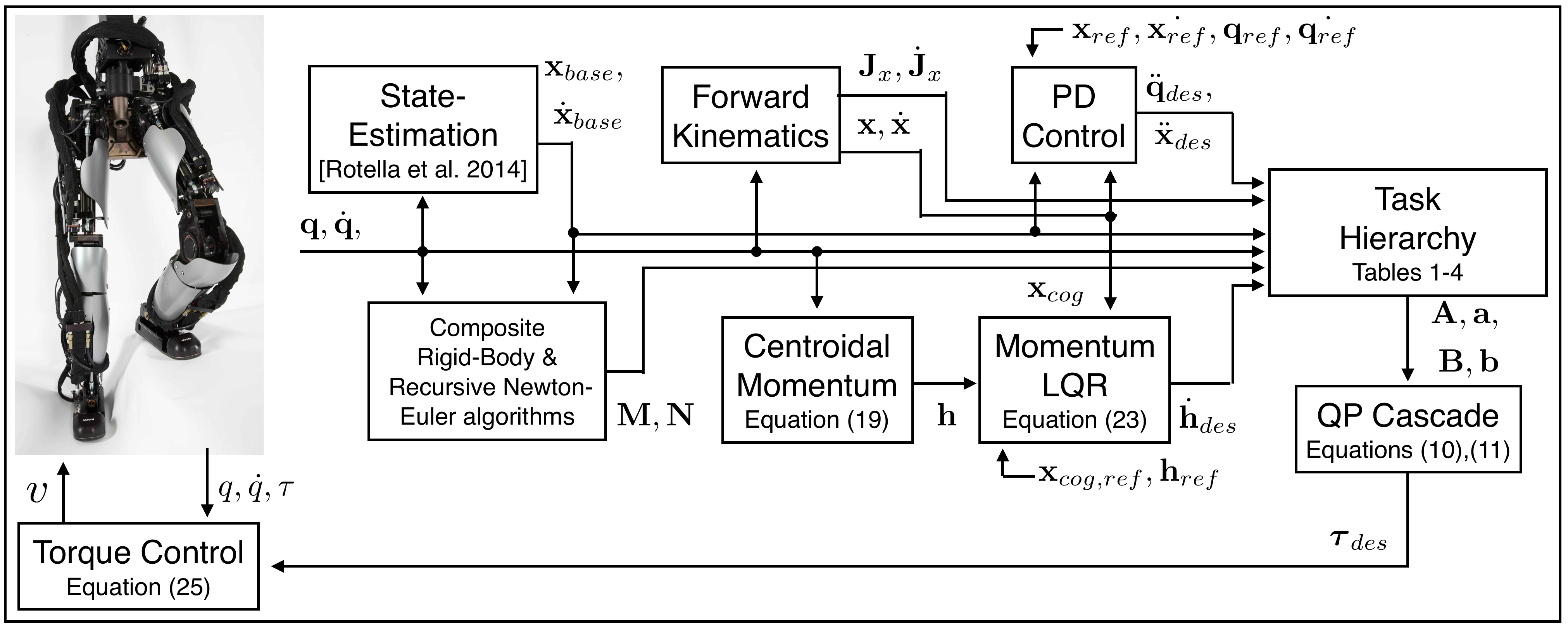}
  \caption{An overview of the control structure used in the presented
    experiments.}\label{fig:control_diagram}
\end{figure*}

We formulated balancing and motion tracking tasks using the algorithm
discussed in Section~\ref{sec:hierarch_inv_dyn} together with the
momentum controller discussed in Section~\ref{sec:lqr_momentum} and
evaluated them on the Sarcos Humanoid described in
Section~\ref{sec:experimental_setup}. The performance of the
controller was evaluated in different scenarios: balancing experiments
and a tracking task in single and double support. A summary of the
experiments is shown in the attached movie \footnote{The movie is also
  available on
  www.youtube.com/ watch?v=jMj3Uv2Q8Xg
}.\\
For all the experiments, we run the hierarchical inverse dynamics
controller as a feedback controller. The desired torque commands
computed by the controller are directly sent to the robot. We do not
use any joint PD controller for stabilization (i.e. feedback control
is only done in task space). A diagram visualizing the flow of control
variables is presented in Figure~\ref{fig:control_diagram}.

\subsection{Processing Time}\label{sec:processing_time}
The computation time of the solver mainly depends on a) the number of
DoFs of the robot b) the number of contact constraints and c) the
composed tasks. All experiments were performed on an Intel Core
i7-2600 CPU with a 3.40GHz processor. Subsequent QPs (cf.
Section~\ref{sec:hierarchical_inv_dyn_solver}) were solved with an
implementation of the Goldfarb-Idnani
dual-method~\cite{Goldfarb:1983ik} using the Eigen matrix library. In
the real robot experiments we use the 14 DoF lower part of a humanoid
to perform several tasks in a \unit[1]{kHz} control loop. In the
following, however, we construct a more complex stepping task in
simulation for the full 25 DoF robot. The goal is to a) evaluate the
speedup from the simplification proposed in Section~\ref{sec:decomp}
and b) give an intuition on how the method scales with the complexity
of the robot.

\begin{table}[thp]
  \center
  \begin{tabularx}{\linewidth}{p{.1\linewidth}p{.28\linewidth}X}
    \hline
    \textbf{Rank} & \textbf{Nr. of eq/ineq constraints}  & \textbf{Constraint/Task} \\
    \hline
    1 &\color{red}{25 eq}&\color{red}{Equation~\eqref{eq:decomp_eq_motion_upper} (not required for simplified problem)}\\
                  &$6$ eq& Newton Euler Equation~\eqref{eq:decomp_eq_motion_lower}\\
                  &$2\times 25$ ineq& torque limits\\
    2 &$c \times 6$ eq& Contact constraints, Eq.~\eqref{eq:contact_constraints}\\
                  &$c\times 4$ ineq& Center of Pressure, Sec.~\ref{sec:task_formulation}\\
                  &$c\times 4$ ineq& Friction cone, Sec.~\ref{sec:task_formulation}\\
                  &$2 \times 25$ ineq& joint acceleration limits, Sec.~\ref{sec:task_formulation}\\
    3 &$3$ eq& PD control on CoG\\
                  &$(2-c)\times 6$& PD control on swing foot\\
    4 &$25+6$ eq& PD control on posture\\
    5 &$c \times 6$ eq& regularizer on GRFs\\
    \hline \hline
                  &\textbf{DoFs:} 25 &\textbf{max. time:}~{\color{red}  \unit[5]{ms}} /  \unit[3]{ms} \bigstrut[t]\\
  \end{tabularx}
  \caption{Full Humanoid Stepping Task for Speed Comparison. The maximum computation time was observed in double support $(c=2)$.}
  \label{tab:simulated_walking}
\end{table}

\begin{figure}
  \centering
  \includegraphics[width=\linewidth]{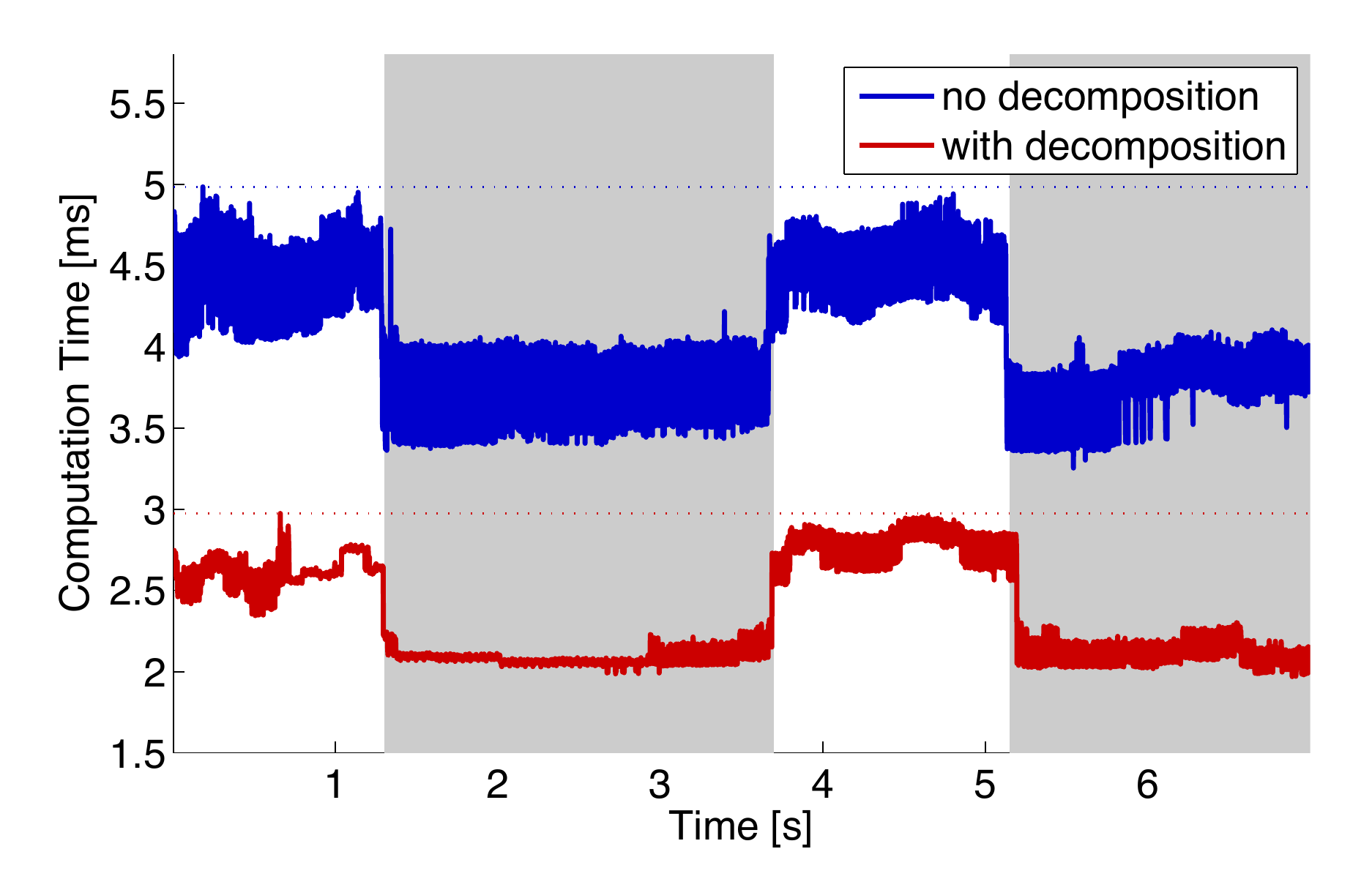}
  \caption{Processing time of a stepping task (see
    Table~\ref{tab:simulated_walking}) using the decomposition
    proposed in Section~\ref{sec:decomp}~(red) and the same task
    performed without the decomposition (blue). The dotted line
    represents the maximum computation per control cycle respectively.
    Intervals shaded in gray show the robot in single support phase.
    In the remaining time the robot is in double support. With the
    proposed decomposition we decreased the computation time by
    approximately 40\%.}
  \label{fig:computation_time}
\end{figure}

We summarized in Table~\ref{tab:simulated_walking} the hierarchy that
is used in simulation. The highest two priorities satisfy hardware
limitations and dynamic constraints, the third priority task tracks a
predefined center of gravity and swing foot motion and the remaining
priorities resolve redundancies on motion and forces. The problem size
changes depending on the number of contacts $c$ ($c=2$ in double
support and $c=1$ in single support). The proposed decomposition
removed 25 equality constraints and 25 optimization variables. We
measured the computation time of both versions of the hierarchical
solver, one with the full EoM and one with the proposed reduction as
plotted in Figure~\ref{fig:computation_time}. Looking at the worst
case (as this is significant for execution in a time critical control
loop) we can reduce computation time by 40\%. In our experiments with
a 14 DoF robot, this speedup allows us to run a \unit[1]{kHz}
control-loop as we will demonstrate in the following sections. It
would not have been possible by using this algorithm without the
simplification. Going from a 14 DoF robot to a 25 DoF robot with
similar task setup makes the peak computation time rise from 1ms to
3ms. In our speed comparison in Figure~\ref{fig:computation_time} one
can see that computation time varies with the number of constrained
end effectors, which can be problematic if the number of contacts
increases too much (e.g. when using both hands and feet).

\subsection{Balance Control Experiments}\label{sec:balance_ctrl}
In the first set of experiments on the robot, we were interested in
systematically evaluating the balance capabilities of the
momentum-based controller with hierarchical inverse dynamics. First,
we compare the performance of the balance control when using the LQR
design and the PD controller described in Section
\ref{sec:lqr_momentum} and then test the performance of the robot when
balancing on a rolling platform and a balancing board.

\subsubsection{Specification of the tasks}
The specification of the task is summarized in Table
\ref{tab:double_sup_pushes} together with the maximum running time for
one control cycle. The physical constraints are put in the highest
priority. In the second priority, we put kinematic contact
constraints, acceleration limits and constraints on reaction forces,
i.e. CoP boundaries and friction cones with a higher weight on CoPs.
In the third hierarchy level, we express our desired closed
loop-dynamics on the momentum together with a PD controller on the
posture and ground reaction force regularization. Here, we prefer to
have the momentum control together with the posture control on the
same level, since the kinematic contact constraints (2nd priority)
lock 12 DoFs and the momentum control another 6 DoFs. Given that we
only have 20 DoFs (including 6 for the floating base), we are left
with too few DoFs to keep a good posture. In our experience this
allowed the robot to keep a better looking posture given the limited
redundancy available.

We did not carefully tune the weights between the tasks in the same
priority level but merely selected an order of magnitude by choosing
among 4 different weights for all tasks, $10^4, 1, 10^{-1}$ or
$10^{-4}$. If not stated otherwise, we chose $\mathbf{P},\mathbf{D}$
gains to be diagonal matrices in all of our tasks except the momentum
control. We put a higher weight on the momentum control since
balancing is our main objective and gave less weight to posture
control and regularization of ground reaction torques.

\begin{table}
  \center
  \begin{tabularx}{\linewidth}{p{.1\linewidth}p{.27\linewidth}X}
    \hline
    \textbf{Rank} & \textbf{Nr. of eq/ineq constraints}  & \textbf{Constraint/Task} \\
    \hline
    1 &$6$ eq& Newton Euler Equation~\eqref{eq:decomp_eq_motion_lower}\\
                  &$2\times 14$ ineq& torque limits\\
    2 &$2 \times 6$ eq& Contact constraints, Eq.~\eqref{eq:contact_constraints}\\
                  &$2\times 4$ ineq& Center of Presure, Sec.~\ref{sec:task_formulation}\\
                  &$2\times 4$ ineq& Friction cone, Sec.~\ref{sec:task_formulation}\\
                  &$2 \times 14$ ineq& joint acceleration limits, Sec.~\ref{sec:task_formulation}\\
    3 &$6$ eq& momentum control, Eqs.~\eqref{eq:lqr_task} or~\eqref{eq:momentum_diag_gain_task}  \\
                  &$14+6$ eq& PD control on posture\\
                  &$2 \times 6$ eq& regularizer on GRFs\\
    \hline \hline
                  &\textbf{DoFs:} 14 &\textbf{max. time:}~ \unit[0.9]{ms} \bigstrut[t]\\
  \end{tabularx}
  \caption{Hierarchy for experiments in Double Support} 
  \label{tab:double_sup_pushes}
\end{table}

\subsubsection{Comparison of momentum controllers}
\begin{table*}[t!]
  \center
  \begin{tabular*}{\textwidth}{p{3.55cm} | c c c c | c c c || c c c c | c c c}
    & \multicolumn{7}{c||}{\hspace{1cm}\textbf{PD Control}} & \multicolumn{7}{c}{\hspace{1cm}\textbf{LQR}} \\
    & \multicolumn{4}{c|}{\textbf{Above Hip Joint}} & \multicolumn{3}{c||}{\textbf{At Hip Joint}} & \multicolumn{4}{c|}{\textbf{Above Hip Joint}} & \multicolumn{3}{c}{\textbf{At Hip Joint}} \\
    & \textbf{F} & \textbf{R} & \textbf{B} & \textbf{L} & \textbf{F} & \textbf{R} & \textbf{L} &  \textbf{F} & \textbf{R} & \textbf{B} & \textbf{L} & \textbf{F} & \textbf{R} & \textbf{L}\\
    \hline
    \textbf{Peak Force} [N] & 233 & 108 & 103 & 80 & 217 & 207 & 202 & 244 & 179 & 107 & 114 & 293 & 223 & 118 \\
    \textbf{Impulse} [Ns] & 7.9 & 4.7 & 5.1 & 3.9 & 9.1 & 9.0 & 6.9 & 6.9 & 6.8 & 6.2 & 4.7 & 9.5 & 8.6 & 4.3 \\
    \textbf{max. CoG Error} [cm] & 4.6 & 3.5 & 3.4 & 2.8 & 5.0 & 4.3 & 2.8 & 3.4 & 2.8 & 2.7 & 1.8 & 3.3 & 3.0 & 1.5 \\
    \textbf{max. Lin. Mom.} [Nm] & 22.6 & 10.2 & 15.8 & 6.9 & 13.1 & 7.9 & 9.3 & 19.8 & 13.6 & 16.3 & 9.1 & 14.4 & 9.2 & 9.5 \\
    \textbf{max. Ang. Mom.} [Nms] & 4.1 & 1.9 & 2.2 & 1.5 & 2.4 & 0.9 & 0.7 & 3.7 & 2.5 & 2.1 & 2.0 & 2.8 & 1.1 & 2.0 \\
  \end{tabular*}
  \caption{Here we list the maximum pushes applied to the robot, where each column shows the properties of one push. When the first 7 pushes were applied, the momentum was controlled with PD control using diagonal gain matrices. For the last seven pushes, LQR gains were used. The robot was pushed from the (F)ront, (R)ight (B)ack and (L)eft either above the hip joint or at the height of the hip joint. The first two rows describe the peak force and impulse of each push. The last 3 rows show the maximum deviation of the CoG  and the linear and angular momentum of the robot after an impact.}
  \label{tab:max_push_stats}
\end{table*}

\begin{figure}
  \centering
  \includegraphics[width=\linewidth]{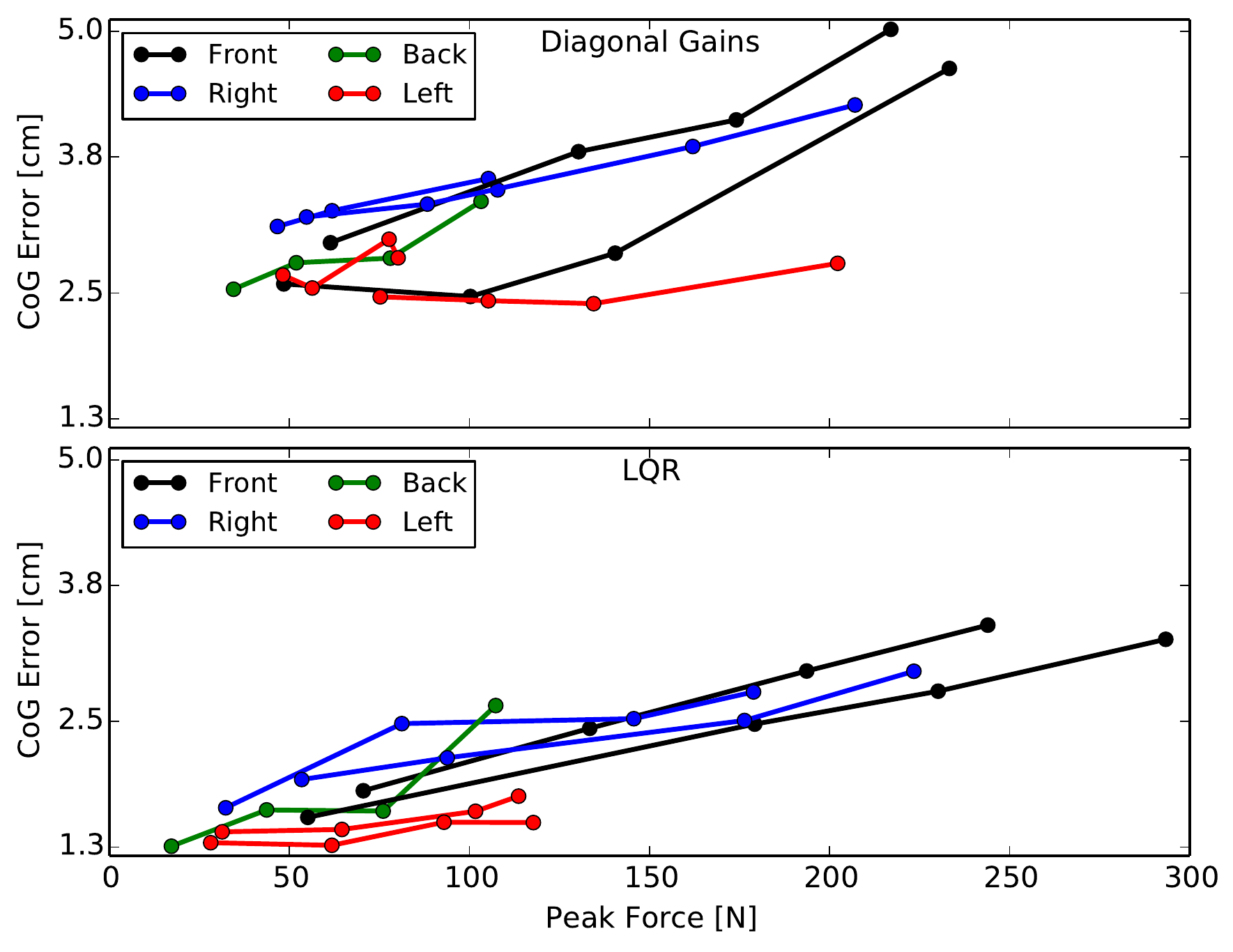}
  \includegraphics[width=\linewidth]{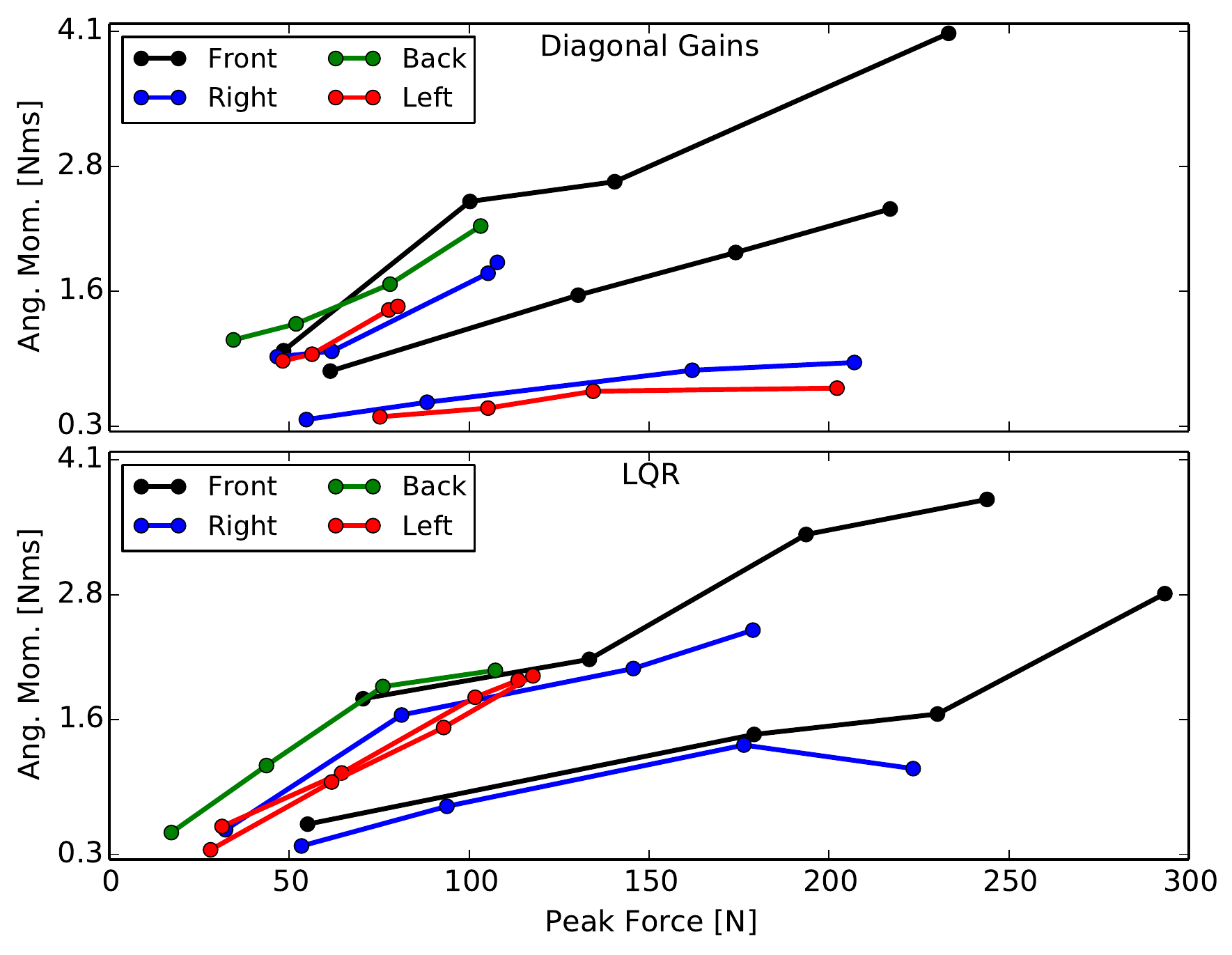}
  \caption{The robot was pushed from 4 sides at the block above the
    hip and at the hip. At each point of attack it was pushed 4 times.
    This figure plots the peak forces of the pushes against the
    maximum CoG displacement (top two figures) and against the maximum
    angular momentum (bottom two figures). The 1st and 3rd figures
    from the top show experiments performed with diagonal gain
    matrices. In the 2nd and 4th plot experiments were conducted with
    the LQR momentum controller. The impulses of the pushes were
    increasing roughly linearly with the peak forces. A list of peak
    impulses is shown in Table~\ref{tab:max_push_stats}. It can be
    seen that overall the CoG error remains lower with the LQR
    controller, while the angular momentum behaves
    similarly.}\label{fig:push_response_stats}
\end{figure}

In this experiment we pushed the robot impulsively with our push-stick
at various contact points with different force directions. The robot
was pushed at 4 points on the torso above the hip (from the front,
right, back and left) and at 3 points at hip level (from the front,
right and left) as can be seen in the attached video. The electronics
of the robot are attached at the back part of the hip which is why we
did not push it at that point. At each of the 7 points we applied 4
pushes of increasing impact up to peak forces of 290 N and impulses of
9.5 Ns, which is on the upper scale of pushes in related
work~\cite{Ott:2011uj,Mason:2014,Pratt:2012ug}. This episode of
experiments is executed with both variations of momentum control
discussed in Section~\ref{sec:lqr_momentum}: PD control with diagonal
gain matrices and with optimal gains from LQR design. We put a
reasonable amount of effort into finding parameters for both
controllers in order to be able to compare them and in both cases we
tried to find parameters that would lead to fast damping of momentum,
with a slight preference for damping linear momentum to ensure that
the CoG was tracked properly. Our resulting LQR performance cost was

\begin{equation}
  \sum_t^{\infty} \begin{bmatrix}\mathbf{x}_{cog} \\ \mathbf{h}\end{bmatrix}^T \mathbf{Q}
  \begin{bmatrix}\mathbf{x}_{cog} \\ \mathbf{h}\end{bmatrix} 
  + \boldsymbol{\lambda}^T \mathbf{R} \boldsymbol{\lambda},
\end{equation}

with $\mathbf{Q}={\it diag}([30, 30, 50, .5, .5, .5, .1, .1, .1]),$
$\mathbf{R}={\it diag}([0.1, 0.1, 0.01, 2, 2, 2])$. For both momentum
control tasks, the robot was able to withstand impacts with high peak
forces and strong impulses without falling. For every push, the change
in momentum was damped out quickly and the CoG was tracked after an
initial disturbance. While it is difficult to compare the performance
of the controllers with other existing state-of-the-art algorithms
because very little quantitative data is available and performance can
drastically change across robots, it seems that both controllers
perform at least as well as, if not better than, other approaches for
which data is available~\cite{Ott:2011uj,Mason:2014,Pratt:2012ug}.
Indeed, the robot was able to absorb impact up to peak forces of 290 N
and impulses of 9.5 Ns. We summarized the information for the
strongest pushes in each direction in Table \ref{tab:max_push_stats}
as a reference.

\begin{figure*}
  \centering \subfigure[Momentum Control with Diagonal Gain
  Matrices]{\begin{minipage}{0.49\textwidth}
      \includegraphics[width=\textwidth]{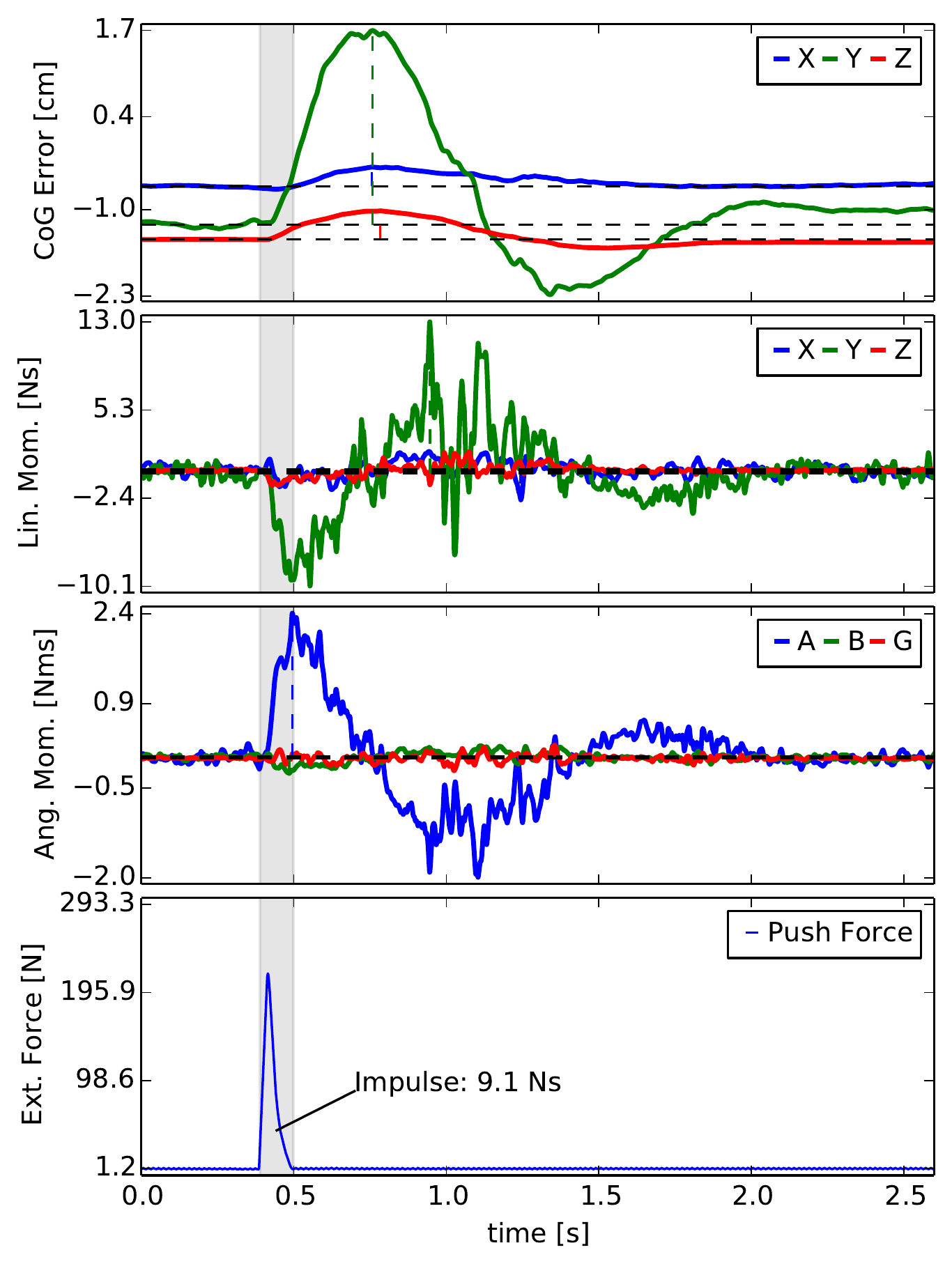}
    \end{minipage}} \subfigure[Momentum Control with LQR
  Gains]{\begin{minipage}{0.49\textwidth}
      \includegraphics[width=\textwidth]{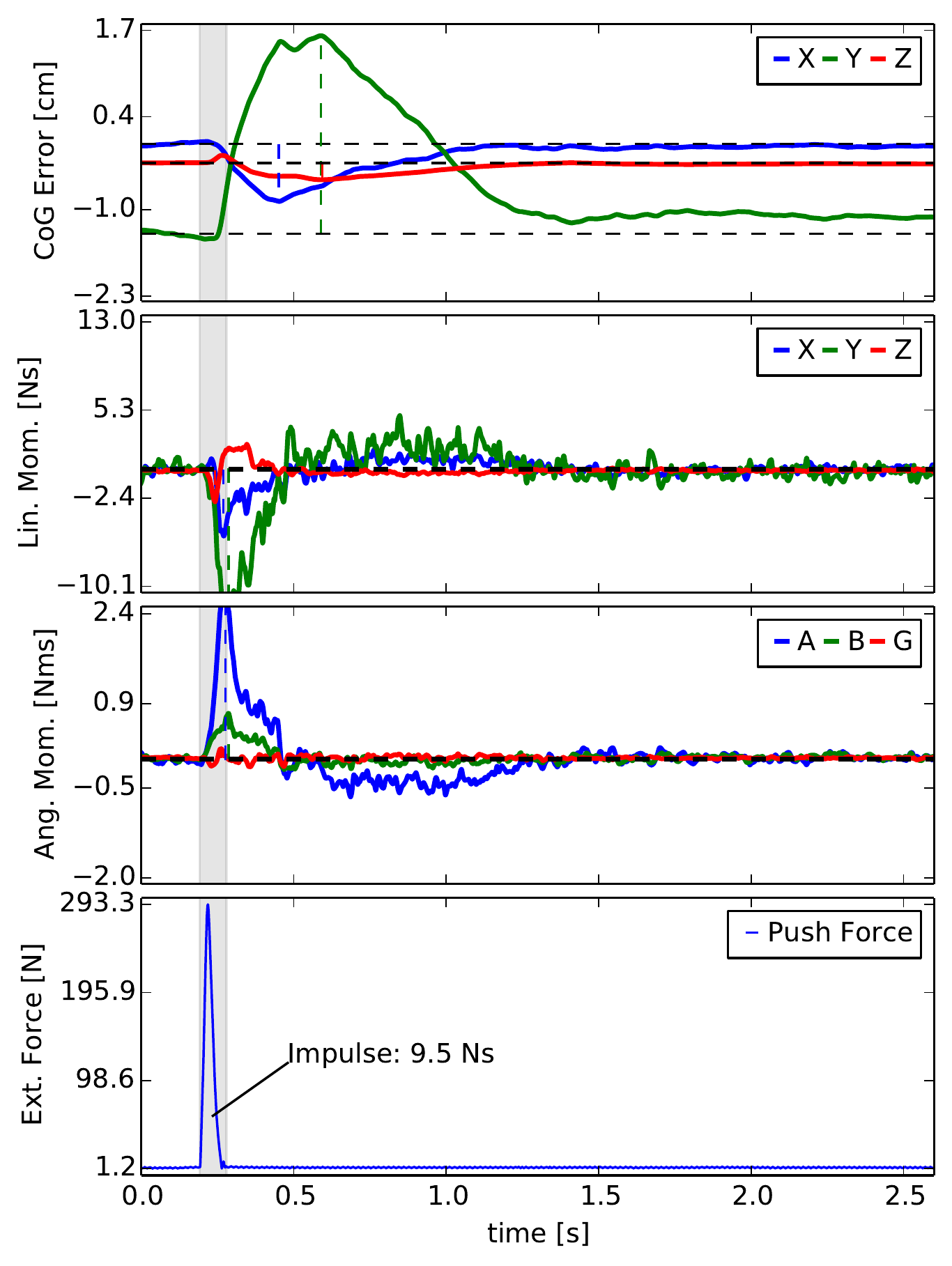}
    \end{minipage}}
  \caption{In this figure we compare typical push recoveries when we
    run momentum control with diagonal PD gain matrices (left) and
    with LQR gains (right). Although the push is stronger for the LQR
    controller (bottom plots), the CoG error (top plots) does not
    deviate from its stationary position more than with the PD
    controller. Both the linear and angular momentum of the robot
    (middle two plots) are damped out quickly by the LQR controller
    and the CoG comes to rest faster as
    well.}\label{fig:lqr_vs_diagonal_gains}
\end{figure*}

In Figure~\ref{fig:push_response_stats} we systematically plotted the
measured peak forces against the maximum deviation of the CoG and
angular momentum for both controllers in order to see the typical
behavior of the robot. The impulses are not plotted as they were
proportional to the peak forces in all our experiments. The maximum
error for both angular momentum and CoG tended to be proportional to
the peak force for all experiments. We notice from the figure that for
both momentum controllers we get similar maximum deviations in angular
momentum. However, with the LQR gains we see a significant improvement
in recovering the CoG. From Figure~\ref{fig:lqr_vs_diagonal_gains} we
also see how the LQR controller recovers quicker although the robot
was pushed harder than with the controller using diagonal gain
matrices.

Figure~\ref{fig:lqr_vs_diagonal_gains} shows a typical response for
both controllers where we plotted the impact force together with the
CoG tracking error and the momentum. We notice that in both cases the
disturbance is damped quickly. We notice that although the peak force
is higher for the LQR controller, the response is better behaved than
for the PD controller and the momentum is damped faster.

While it is always difficult to ensure that a better set of parameters
couldn't be found for the PD controller, this result suggests that the
LQR design performs better than the PD controller. Moreover, the LQR
design is much simpler to tune because the design of a performance
cost has a more intuitive meaning than PD gains and it can capture the
coupling between linear and angular momentum. The other advantage of
the LQR design is that once the cost function is fixed, new gains can
be computed for various poses and desired momentum behaviors
automatically without manual re-tuning. This aspect was very helpful
for the contact switching and single support task that we present in
the following section.

The balance controller that is implemented does not rely on co-planar
feet and is able to produce very compliant behaviors. When we pushed
the robot with a constant force at various parts, it stayed in balance
and adapted its posture in a compliant manner. We also tested the
controller when the feet were not co-planar, but one foot was put on
top of a block. These behaviors can be seen in the attached movie.

\subsubsection{Balancing on moving platforms}
\begin{figure}
  \centering
  \includegraphics[width=\linewidth]{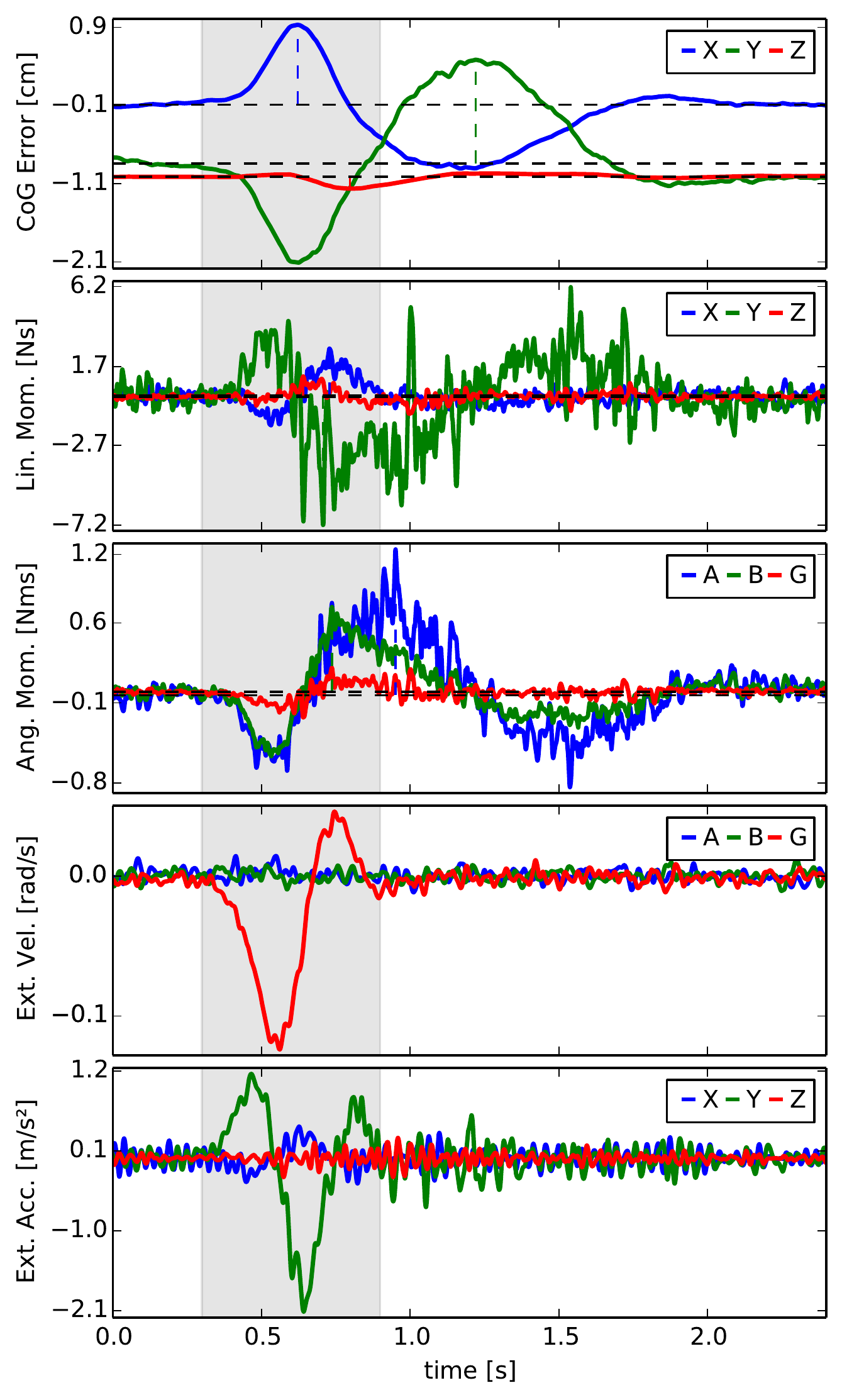}\\
  \includegraphics[width=.7\linewidth]{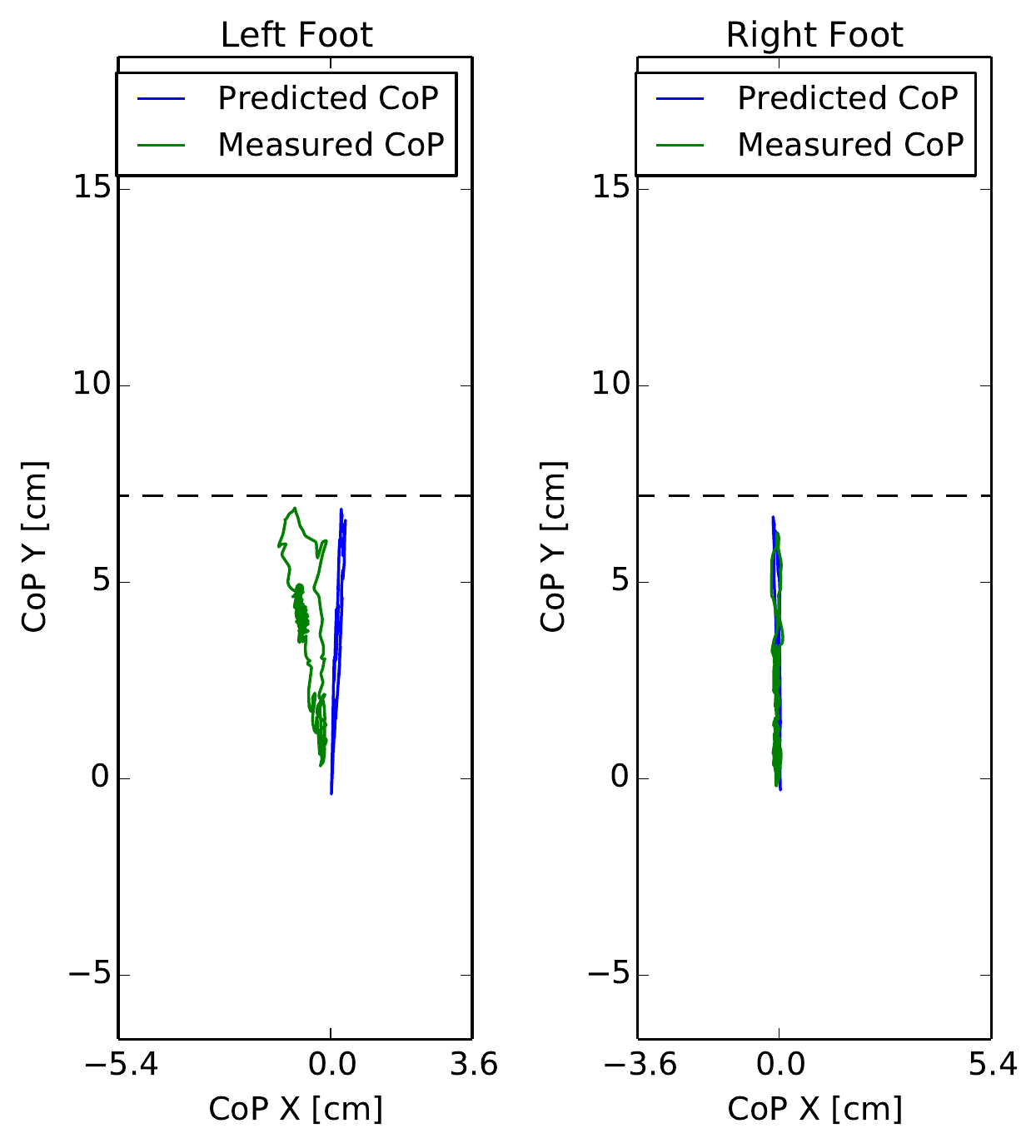}
  \caption{The top three plots show the CoG error and momentum when
    the robot is balancing on the rolling platform. The next two
    figures plot the angular velocity and linear acceleration of the
    platform. The platform is displaced in Y direction with the robot
    facing the direction of the disturbance. The bottom figures show
    predicted and measured CoPs. Please refer to the text for a
    discussion of the results.}
  \label{fig:rolling_platform}
\end{figure}

For our next experiment, we put the biped on a rolling platform and
rotated and moved it with a rather fast change of directions. A
typical behavior of the robot is plotted in
Figure~\ref{fig:rolling_platform}. The angular velocity and linear
acceleration measured from the IMU that we attached to the rolling
platform are plotted together with the CoG tracking error and
momentum. Although the CoG is moving away from its desired position
when the platform is moving around, it remained bounded, the momentum
was damped out and the robot kept standing and recovered CoG tracking.
The stationary feet indicated that forces were applied that were
consistent with our CoP boundaries. We notice from the figure that the
CoPs, as they were predicted from the dynamics model, are
approximately correct. However, one can expect that a higher precision
might be needed to achieve dynamic motions which could be achieved
with an inertial parameters estimation procedure \cite{Mistry:2009dh}.

\begin{figure}
  \centering
  \includegraphics[width=\linewidth]{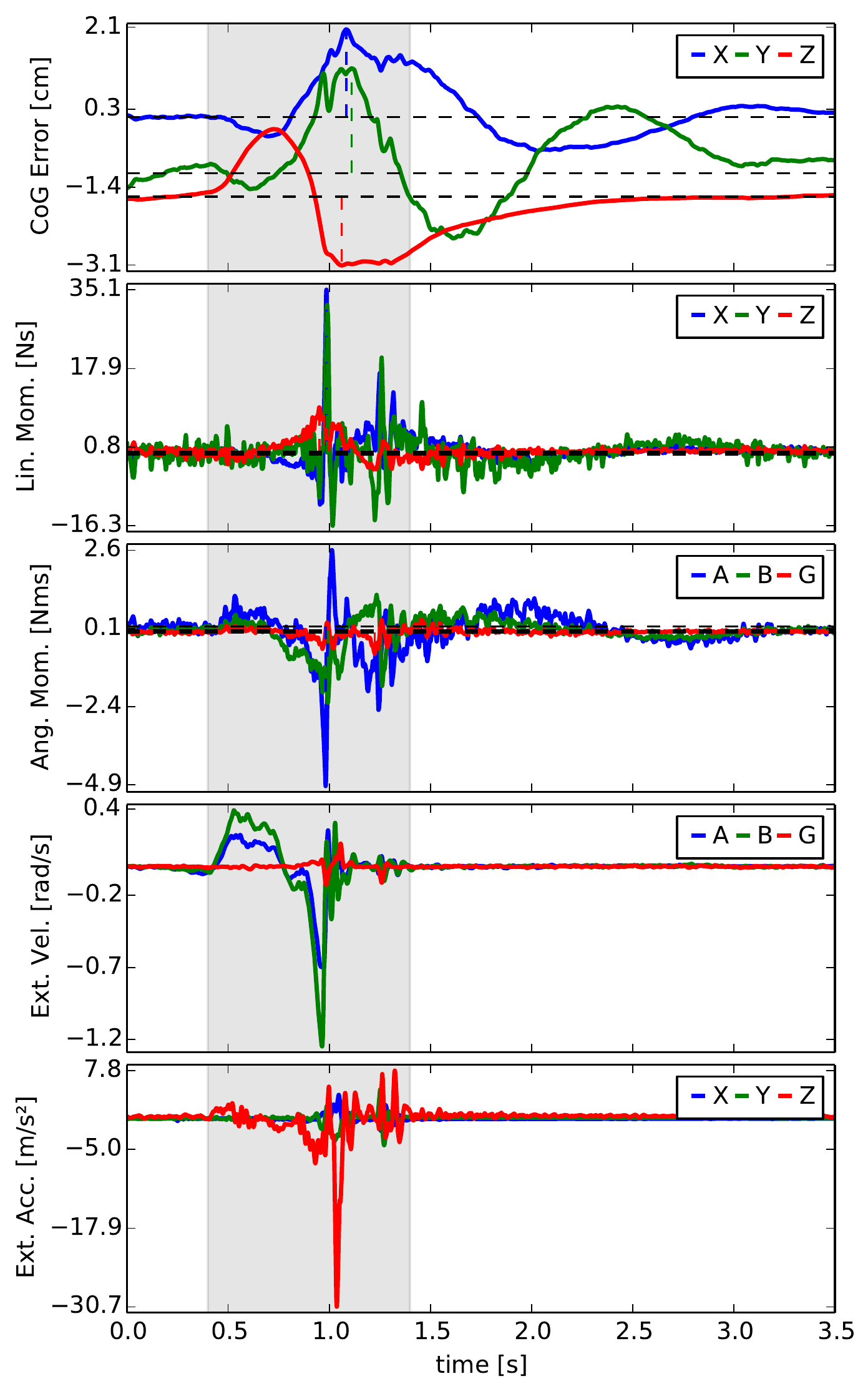}\\
  \includegraphics[width=.7\linewidth]{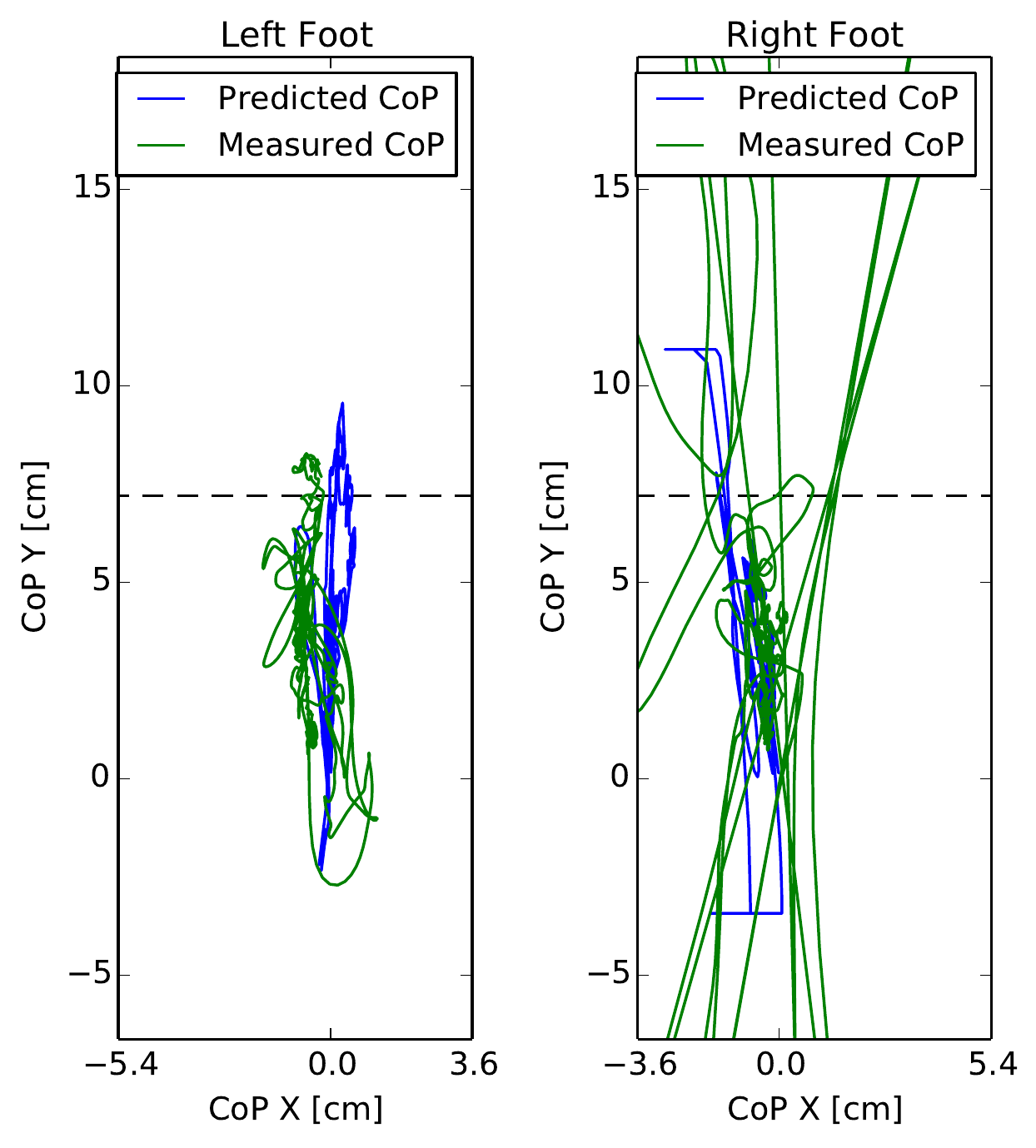}
  \caption{The top figure plots the robot CoG error and momentum when
    it was dropped from the highest point on the balancing board. In
    the bottom two plots one can see the angular velocity and linear
    acceleration of the balancing board, where we can identify the
    moment of impact with the ground at \unit[1]{s}. At that moment
    the right foot bounced off the ground and lost contact for an
    instance of time as can be seen in the measured CoP in the bottom
    plots and in the video. The admissible CoP hit the boundary and
    saturated in Y direction. Still, the robot was able to stabilize
    its feet and CoG and damp out the momentum.}
  \label{fig:seesaw}
\end{figure}

In an additional scenario, the biped was standing on a balancing
board. We ran the experiment with two configurations for the robot: in
one case the robot is standing such that the board motion happens in
the sagittal plane and in the other case the motion happens in the
lateral plane. In this case, the slope was varied in a range of
$[-2.8^\circ; 5.6^\circ]$ elevating the robot up to \unit[6.9]{cm}.
Even for quite rapid changes in the slope, the feet remained flat on
the ground. Compared to the push experiment the CoPs were moving in a
wider range, but still remained in the interior of the foot soles with
a margin. In this case, we notice a discrepancy in the predicted
contact forces and the real ones, making the case for the need for a
better dynamics model. Eventually, we dropped the robot from the
maximum height (Figure~\ref{fig:seesaw}), such that the feet bounced
off the ground at impact and tilted for a moment as can bee seen in
the video. Yet, the robot recovered and was still able to balance. We
notice in the figure that in this case, the measured CoP of the foot
that was lifted drastically differs from the predicted ones. We also
note that the predicted CoPs reach their admissible boundary as is
seen from the flat horizontal lines.

When we increased the pushes on the robot, eventually the momentum
could not be damped out fast enough anymore and the robot reached a
situation where the optimization could not find solutions that would
balance the robot anymore (i.e. the slack variable associated to the
desired momentum becomes very high) and the biped fell. In these cases
the constraint that the feet have to be stationary was too
restrictive. A higher level controller that takes into account
stepping (e.g. \cite{Pratt:2012ug,Urata:2012fb}) becomes necessary to
increase the stability margin.

\subsection{Tracking Experiments in Double Support}
\begin{figure}
  \centering
  \includegraphics[width=\linewidth]{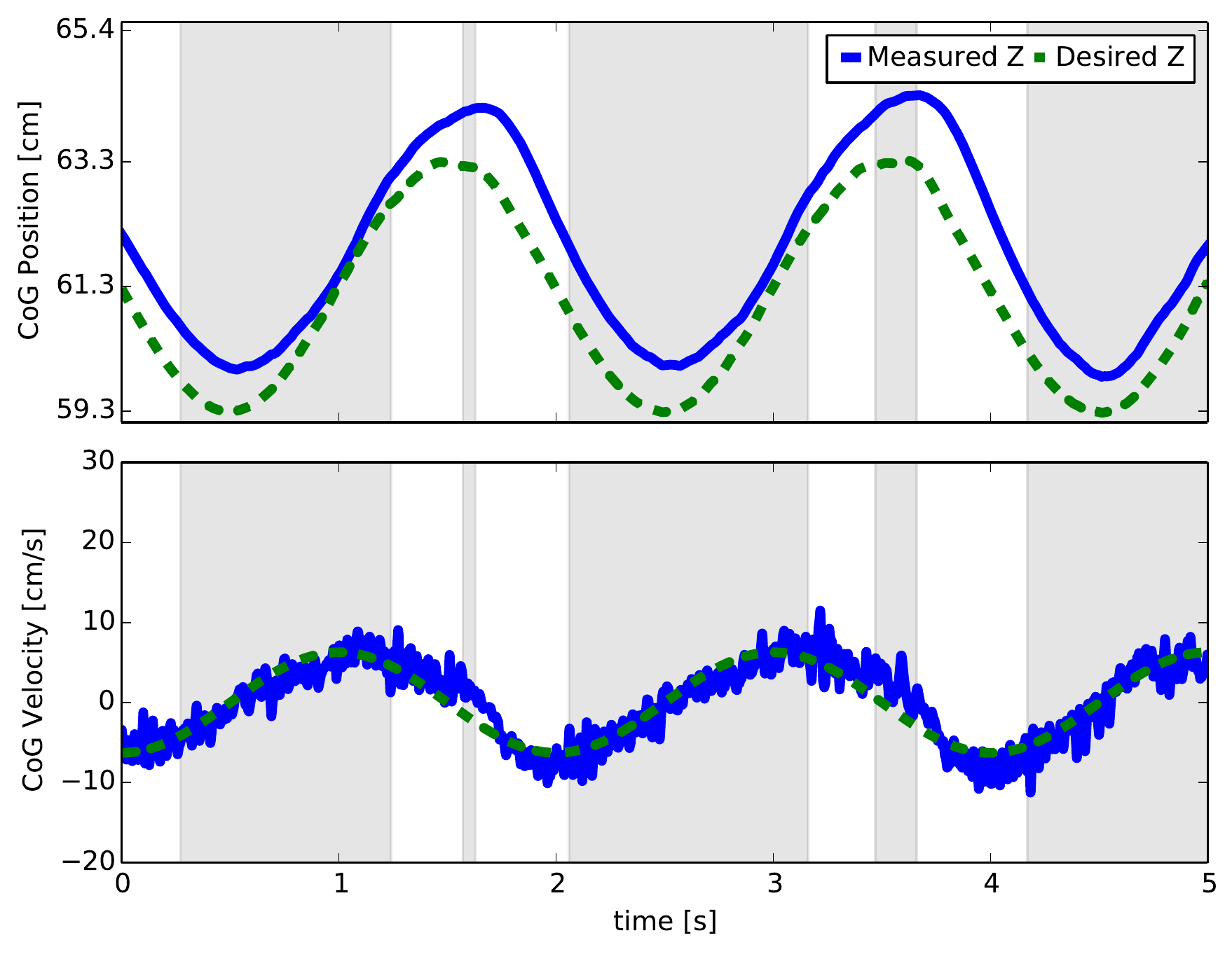}\\
  \includegraphics[width=.7\linewidth]{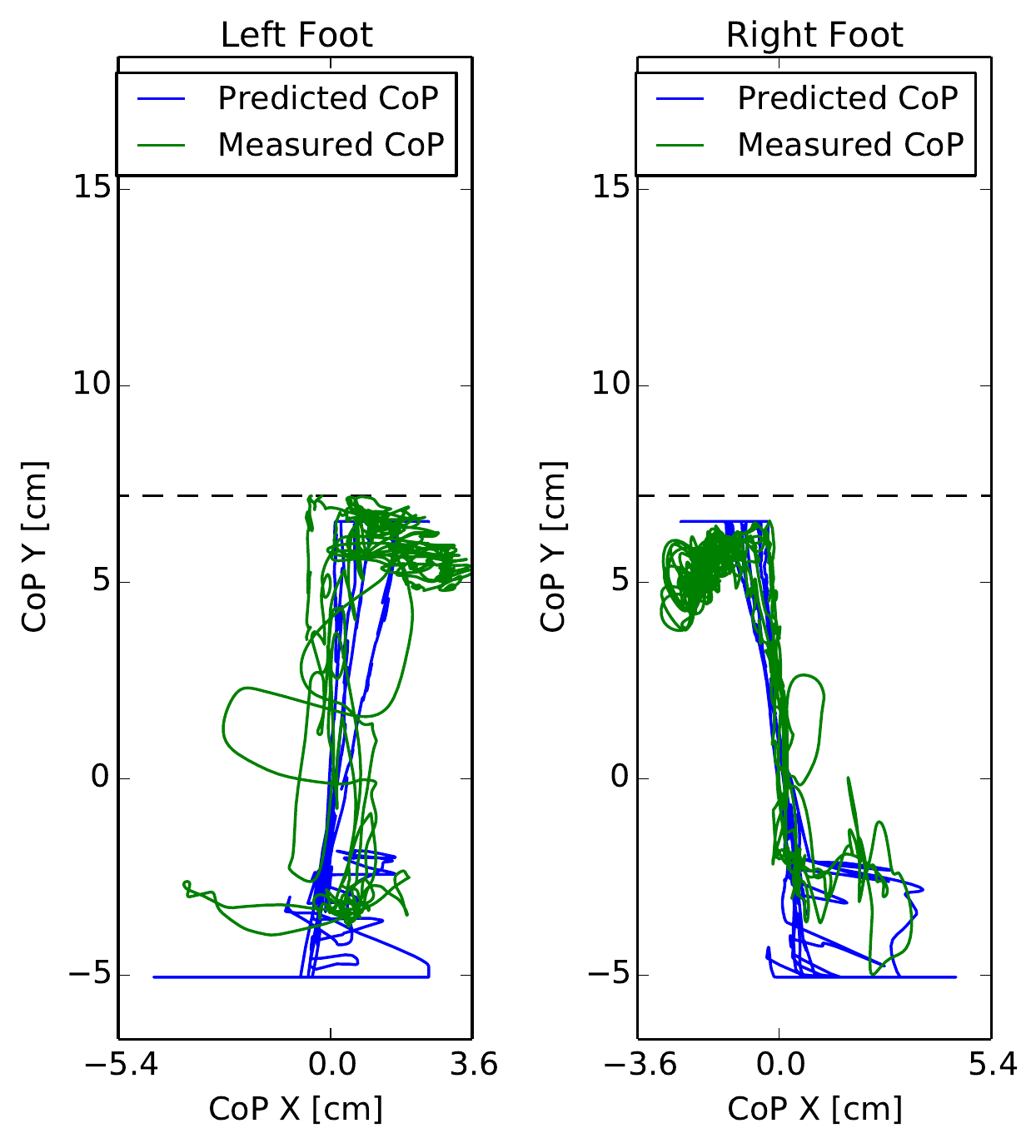}
  \caption{Tracking of the CoG in vertical direction during the
    squatting task when CoP constraint is active. The top two plots
    show the CoG desired and actual vertical positions and velocities.
    The grayed area corresponds to periods during which the CoP
    constraint is active. The bottom plot shows the evolution of the
    CoPs for both feet. The horizontal blue lines of the desired CoP
    correspond to an active constraint.}
  \label{fig:squatting}
\end{figure}

In the next experiment the robot is performing a squatting like
motion. We keep the same task hierarchy as in the balance experiments
(see Table~\ref{tab:double_sup_pushes}) and make the CoG track sine
curves of \unit[0.3]{Hz} and \unit[0.5]{Hz}. The CoG is moving with an
amplitude of \unit[2]{cm} in the forward direction and an amplitude of
\unit[3]{cm} in the vertical direction. The results can be seen in the
attached video.

In order to demonstrate the utility of the hierarchy, we setup a
squatting experiment such that CoP constraints would be activated
during the motion. We show the result of this experiment in
Figure~\ref{fig:squatting}. We notice that the CoP constraints are
active in most parts of the experiments. This constraint prevents the
feet from tilting and the robot stays stable. We also notice that the
real CoPs follow the predicted ones relatively well and stay inside
the support polygon. As a consequence, the tracking of the CoG, which
is in a lower priority, is not ideal but still achieves a reasonable
performance. CoG velocity tracking is still achieved reasonably well.
We note that the discrete activation/deactivation of the constraint is
not directly visible on the CoG motion behavior.

This experiment illustrates the importance of hierarchies. By solving
a QP without a hierarchy, there would be two possibilities, either the
CoP constraint is set as a hard constraint of the optimization problem
and there is no guarantees that a solution to the problem exists or it
is put in the cost function with the CoG task and the solution is
necessarily a trade-off between contact constraints and motion tasks.
Exploiting a hierarchical setup, we are guaranteed to find a solution
to the optimization problem and at the same time we are guaranteed
that the CoG tracking task does not interfere with the CoP constraint.

\subsection{Single Support Experiments}
The experiments in the previous sections were done while the robot
remained in double support. The goal of this experiment is to show
that the controller can handle more complicated tasks involving
contact switching and that the robot is able to balance on a single
foot in face of disturbances. Further, we evaluate all capabilities in
a single task: contact switching, push rejection in single support and
tracking a leg motion.

First the robot moves from a double support position to a single
support position where the swing foot is lifted \unit[10]{cm} above
the ground while balancing. This motion consists of 3 phases. First,
the robot moves its CoG towards the center of the stance foot. Then an
unloading phase occurs during which the contact force regularization
enforces a zero contact force to guarantee a continuous transition
when the double support constraint is removed. In the final phase, a
task controlling the motion of the swing foot is added to the
hierarchy. Our contact switching strategy is simple but guarantees
that continuous control commands were sent to the robot. For more
complicated tasks, such guarantees can always be met by using
automatic task transitions such as in \cite{jarquin:humanoids:2013}.
The composition of hierarchies is summarized in
Table~\ref{tab:one_foot}. Concerning computation time, the controller
computes a solution in average well below 1ms but a maximum at 1.05ms
is reached a few times during the unloading phase due to many
constraints becoming active at the same time.

Once in single support, we pushed the robot to verify that it is
balancing. Impacts were applied with peak forces up to \unit[150]{N}
and impulses between \unit[4.5]{Ns} and \unit[5.8]{Ns}. We saw a quick
recovery of the CoG while the CoPs stayed bounded. In order to control
the foot we used Cartesian position control (i.e. the swing foot task
consists of a PD controller for the foot position in Cartesian space).

\begin{table}
  \center
  \begin{tabularx}{\linewidth}{p{.1\linewidth}p{.28\linewidth}X}
    \hline
    \textbf{Rank} & \textbf{Nr. of eq/ineq constraints}  & \textbf{Constraint/Task} \\
    \hline
    1 &$6$ eq& Newton Euler Equation~\eqref{eq:decomp_eq_motion_lower}\\
                  &$2\times 14$ ineq& torque limits\\
    2 &$2\times 4$ ineq& Center of Pressure, Sec.~\ref{sec:task_formulation}\\
                  &$2\times 4$ ineq& Friction cone, Sec.~\ref{sec:task_formulation}\\
                  &$2 \times 14$ ineq& joint acceleration limits, Sec.~\ref{sec:task_formulation}\\
    3 &$6$ eq. & Linear and angular momen-\\&&tum control\\
                  &$12 / 6$ eq.& Contact constraints, Eq.~\eqref{eq:contact_constraints} \\
                  &$0 / 6$ eq. & Foot motion (swing)\\
                  &$14$ eq.& PD control on posture\\
    4 &$2 \times 6 / 1 \times 6$ eq.& regularizer on GRFs\\
    \hline \hline
                  &\textbf{DoFs:} 14 &\textbf{max. time:}~ \unit[1.05]{ms} \bigstrut[t]\\
  \end{tabularx}
  \caption{Hierarchy for Single Support  experiments}
  \label{tab:one_foot}
\end{table}

\begin{figure}
  \centering
  \includegraphics[width=\linewidth]{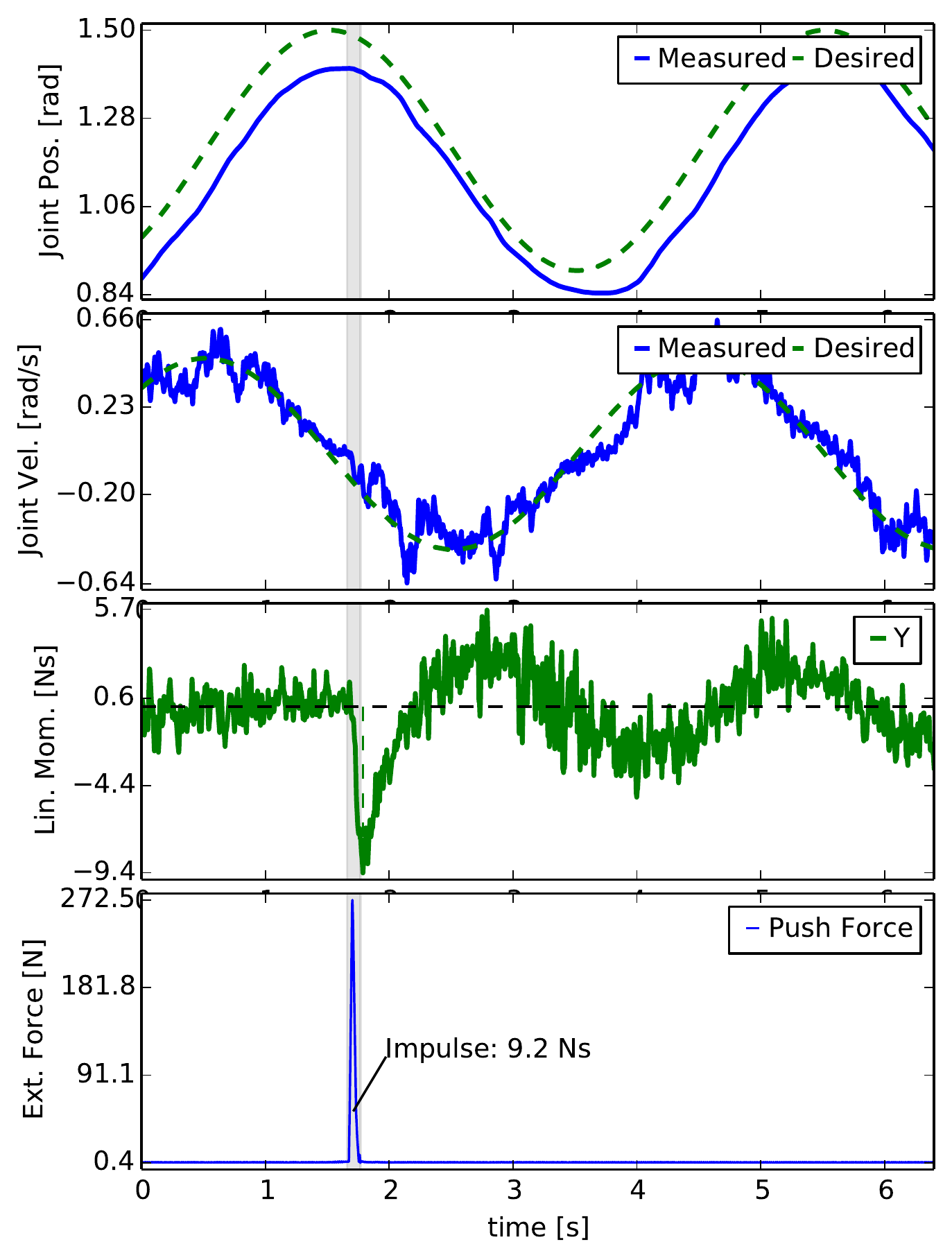}
  \includegraphics[width=.35\linewidth]{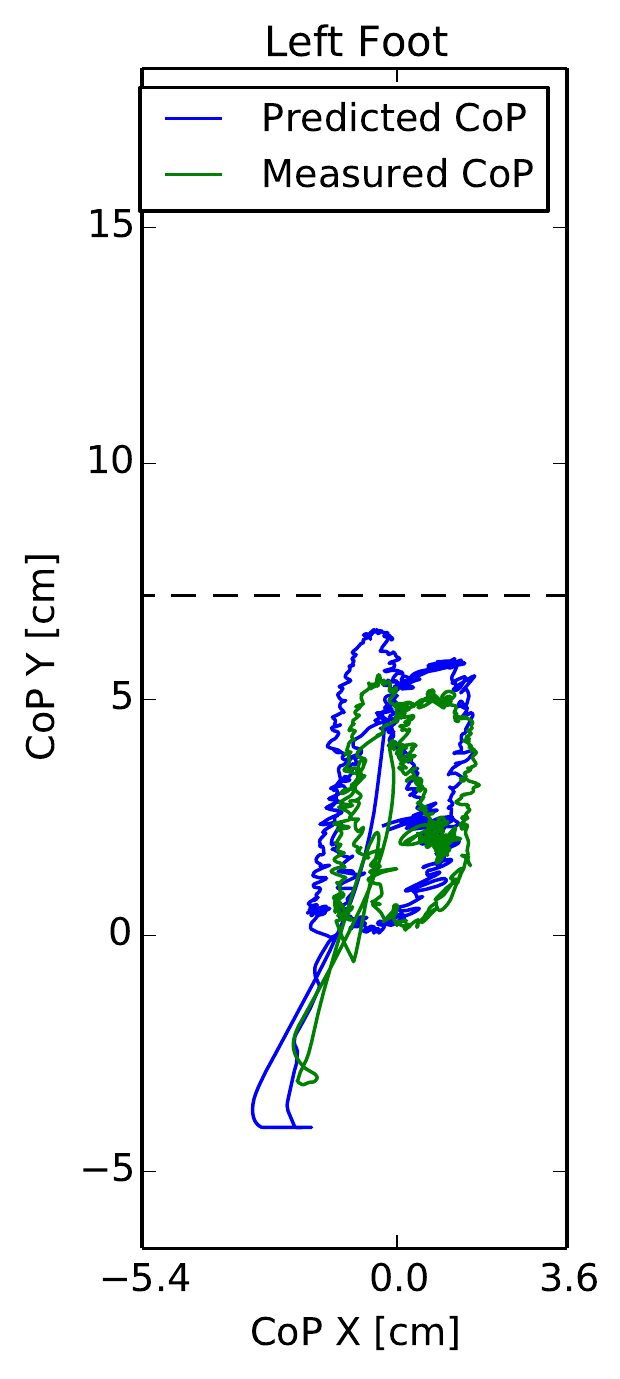}
  \caption{Tracking error on the position and velocity of the knee
    joint (top two plots) when the robot is standing on one leg and
    pushed with a peak force of \unit[270]{N} (bottom plot). Although
    the linear momentum of the system (3rd plot from top) was
    perturbed for a moment, joint tracking was barely affected. The
    bottom plot shows the CoP of the stance foot, which saturates
    close to the heel during the push, such that the foot does not
    start tilting.}
  \label{fig:single_support_pushes}
\end{figure}

We repeat the experiment where the robot, once it is standing on one
leg, swings its leg up and down tracking \unit[0.25]{Hz} sine curves
on the hip and knee flexion/extension joints (as can be seen in
Figure~\ref{fig:single_support_pushes}). In this case, we swap the
Cartesian foot control for a desired trajectory of the hip and knee in
joint space (i.e. a task that consists of time-varying positions for
both joints). Indeed, while in simulation Cartesian tracking is
perfect, on the real robot the tracking performance of the Cartesian
task of the swing foot is not satisfactory when moving at higher
speeds and amplitudes. We suspect that several reasons can explain the
problem. One of the reasons seems to be due to the amount of noise
present in the position sensors such that it is difficult to increase
the position gains while still being able to damp the system. Since
the feedback loop is closed around the foot position, which is
estimated through forward kinematics, its velocity estimation seems to
suffer from the cumulative noise of all the sensors. In this case, a
direct joint control suffers less from that problem. Another problem
could come from an insufficiently accurate model of the swing leg
dynamics where unmodeled dynamics could become more dominant.

While the robot is performing the task, it is pushed strongly at the
hip from the front as can be seen in the video. The joint tracking
together with the linear momentum and the push force are shown in
Figure~\ref{fig:single_support_pushes}. A spike in momentum can be
seen at the moment of impact, but is damped and remains bounded.
During the push, the CoP constraint is activated when the CoP comes
close to the heel. Thanks to the inequality constraint, the foot does
not start tilting and the robot can recover from the push. What is
remarkable is that the swing leg tracking is barely affected although
the push is comparable to the strongest impacts we applied in double
support. It is worth mentioning again that the foot size of the robot
is rather small compared to other humanoids.

\section{Discussion}\label{sec:discussion}
In the following we discuss the results we presented and how they
relate to other approaches.

\subsection{Task design and hierarchies}
In our experiments, we exploit hierarchical separation of physical
constraints and tasks, such as the robot dynamics and reaction force
constraints and balancing tasks. This is guaranteed to always find a
feasible solution while generating physically consistent solutions and
generating admissible task dynamics as close as possible to the
desired ones. As we have shown in the experiments, this is important
to keep balance in cases when reaction force constraints conflict with
lower priority tasks. Here, hierarchies can guarantee that performance
in balancing is traded off, but the physical consistency of the
solution is always guaranteed. Note that with a QP formulation this
would not be possible.

From our experience, we prefer to keep posture tracking task and
momentum control in the same priority and adapt importance by weights.
It seems that the 14 DoFs of our robot are too limiting and do not
leave enough freedom when posture control is put in a separate rank.
Using a full humanoid with arms will increase the flexibility of the
robot and allow for more hierarchy levels. We expect to have, for
example, manipulation tasks in a higher priority than balancing tasks.

As mentioned in the experiment description we used only a small set of
weights for our tasks. This already gives us a balancing performance
that is at least as good as related work if not better. Better
performance can be achieved by adjusting parameters and hierarchical
setup more specific to tasks of interest~\cite{herzog:2014b}. It is
important to note that in both contributions we do not use joint PD
stabilization, but verify that the performance is solely the effect of
the hierarchical inverse dynamics controller.

\subsection{Relation to other balancing approaches}
The balance control strategy used in this paper is similar to the
formulation of the momentum-based controller presented by Lee and
Goswami \cite{Lee:2012hb,Lee:2010em}. Our formulation has the great
advantage of solving a single optimization problem instead of several
ones and can therefore guarantee that the control law will be
consistent with all the constraints (joint limits, accelerations,
tor\-que saturation, CoP limits and contact force limitations). As we
have seen in the experiments, consistency with inequality constraints
can be very important to improve robustness. Furthermore, we search
over the full set of possible solutions and thus we are guaranteed to
find the optimum, where~\cite{Lee:2012hb,Lee:2010em} optimize over
sub-parts of the variables sequentially. It is also different from the
work of \cite{Kajita:2003gj} because we explicitly take into account
contact forces in the optimization and not purely kinematics, which
allows us to optimize the internal forces created by the contacts.

The balance controller is related to the work of
\cite{Stephens:2010vu}. In \cite{Stephens:2010vu}, the authors write
the whole optimization procedure using
Equation~\eqref{eq:equations_of_motion} with constraints similar to
the ones we use. However, with the optimization problem being
complicated, they actually solve a simpler problem where the contact
forces are first determined and then desired accelerations and torques
are computed through a least-square solution. From that point of view,
torque saturation and limits on accelerations are not accounted for.
In our experiments, no tradeoff is necessary and we solve for all the
constraints exactly. Further, the capability of strict task
prioritization makes the design of more complicated tasks like
balancing on one foot easier. Also, separating the EoM from kinematic
contact constraints allows to keep solutions consistent with the
dynamics even in postures where the feet cannot be kept stationary.

We have also shown in the paper that the use of a LQR design for the
momentum task can simplify the controller design and improve robot
performance by taking into account the coupling between linear and
angular momentum. This design was particularly useful for the contact
switching and single support task. Indeed, using the PD control
approach, it was not possible to use the same gains in double or
single support to regulate the CoG. With the LQR design, the gains for
the momentum control were automatically computed using the same cost
function and therefore no specific gain tuning was needed.

\subsection{Relations to other hierarchical inverse dynamics solvers}

The implemented QP cascade is a combination of the two
algorithms~\cite{Kanoun:2011ey,deLasa:2010hf}.
We use a surjective Nullspace map $\mathbf{Z}_r$ (cf.
Equation~\ref{eq:optimals_constr_1}), similar to~\cite{deLasa:2010hf}.
However, in~\cite{deLasa:2010hf} inequality constraints are included
only in the first priority, i.e.
$\mathbf{A}_1 = \dots = \mathbf{A}_r$. Instead, the proposed method
allows for prioritization among inequality constraints as it was done
in ~\cite{Kanoun:2011ey}. This variant of QP cascades combines the
benefits of both algorithms. On the one hand variables are eliminated
from one QP to the other, resulting in faster solvable QPs. On the
other hand, it allows for prioritization of inequality constraints,
which we exploit e.g. to give more
importance to hardware limitations than to contact constraints.\\
Although fast enough for our experiments on the lower part of our
humanoid robot, the speed comparison in
Section~\ref{sec:processing_time} shows that a more efficient
algorithm is required when we run feedback control on the full 25 DoF
robot.
A method that is tailored to solve inequality constrained hierarchical
tasks was derived in~\cite{Escande:2014en}. With an active set method
dedicated to solve prioritized inequality constraints, it can
outperform QP cascades in terms of speed.
The QP cascade used in this algorithm trades off computational
efficiency to a simpler implementation, where the handling of
inequality constraints is passed on to an off-the shelf QP solver. As
the focus of this paper was the experimental evaluation of the problem
formulation, a QP cascade with the employed modifications was
sufficiently fast and relatively easy to implement.
Another practical advantage of QP cascades is the easily implemented
regularization term in Equation~\eqref{eq:single_qp}, which increases
numerical robustness in face of task singularities as discussed
in~\cite{Kanoun:2011ey}.
The approach of \cite{Escande:2014en} might also be interesting
because it would directly allow the use of warm-start techniques to
speed up the computations. Warm starting the optimizer should
significantly improve computation time since during most tasks the
active set does not change much from one control step to the other.
Any inverse dynamics approach using either QP
cascades~\cite{Mansard:2012gy},\cite{Saab:2013vr} or Hierarchical QP
algorithm might profit from the decomposition proposed in
Section~\ref{sec:decomp}, as it is not required to compute an SVD of
the full equations of motion, but only of the last six rows.

\section{Conclusion}
In this paper, we have presented experimental results using a
hierarchical inverse dynamics controller. A variant of cascades of QPs
was presented and implemented in a \unit[1]{kHz} feedback-control
loop. We used LQR to formulate momentum controllers for balancing and
tracking tasks. Our main focus then was the experimental evaluation of
the control framework on a torque controlled humanoid robot. In a
series of experiments, we evaluated systematically the balancing and
tracking capabilities of the robot. The humanoid showed a robust
performance in single and double support and was able to recover from
pushes and other disturbances. Our results suggest that the use of
complete dynamic models and hierarchies of tasks for feedback control
is a feasible approach, despite the model inaccuracies and
computational complexity. For future work, we would like to integrate
higher level planners to compose more complex tasks such as walking.

\begin{acknowledgements}
  We would like to thank Ambarish Gos\-wami and Seungkook Yun for
  hosting us at the Honda Research Institute for one week and for
  their precious help in understanding the original momentum-based
  controller. We would also like to thank Ambarish Goswami and
  Sung-Hee Lee for giving us an early access to their publication.
  We are also grateful to Daniel Kappler for helping us with the 
  videos. Last, but not least, we would like to thank the anonymous 
  reviewers for their very valuable comments that helped improve 
  the final version of the paper.\\
  This research was supported in part by National Science Foundation
  grants IIS-1205249, IIS-1017134, CNS-0960061, EECS-0926052, the
  DARPA program on Autonomous Robotic Manipulation, the Office of
  Naval Research, the Okawa Foundation, and the Max-Planck-Society.
  Any opinions, findings, and conclusions or recommendations expressed
  in this material are those of the author(s) and do not necessarily
  reflect the views of the funding organizations.
\end{acknowledgements}

\bibliographystyle{spmpsci} 
\bibliography{AURO_2014}

\begin{wrapfigure}[13]{l}{3.6cm}
\includegraphics[width=4cm]{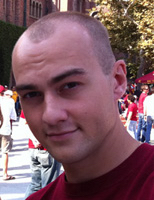}
\end{wrapfigure}
\textbf{Alexander Herzog} is a PhD student at the Max Planck Institute for
Intelligent Systems, T\"ubingen. He
studied Computer-Science at the  Karlsruhe Institute
of Technology, in Germany. Alexander visited the Computational Learning and Motor Control Lab (University of Southern
California) in 2011,
where he worked on the problem of grasp planning for his diploma
thesis. After receiving his Diploma in the same year, he joined
the Autonomous Motion Laboratory at the  Max-Planck Institute for
Intelligent Systems in 2012. His research interests include contact
interaction in whole-body control and grasping for humanoids.\\

\begin{wrapfigure}[13]{l}{3.6cm}
\vspace{-.75cm}
\includegraphics[width=4cm]{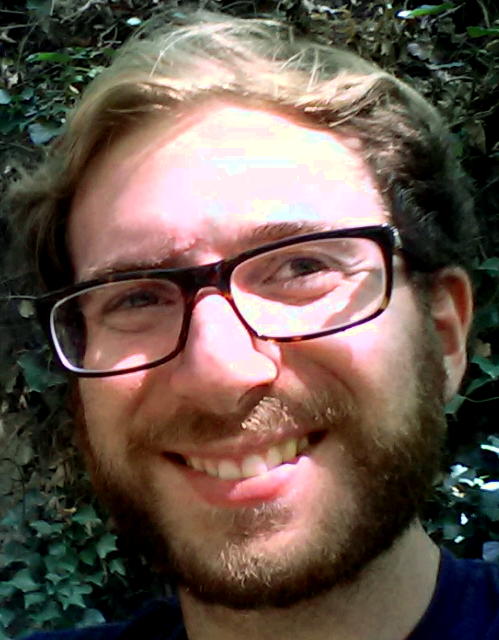}
\end{wrapfigure}
\textbf{Nicholas Rotella} studied Mechanical Engineering at The Cooper
Union for the Advancement of Science and Art in New York, NY from 2008
to 2012, after which he joined the CLMC lab at USC to pursue a PhD in
Computer Science. He passed his PhD screening process and received his
Masters degree in Computer Science from USC in 2014 and continues to
work towards his PhD.  His research interests include optimal
estimation and control for legged robotic systems, controlling contact
interactions for locomotion and manipulation, trajectory optimization
and planning of complex motor control tasks.\\

\begin{wrapfigure}[9]{l}{3.6cm}
\vspace{-.75cm}
\includegraphics[width=4cm]{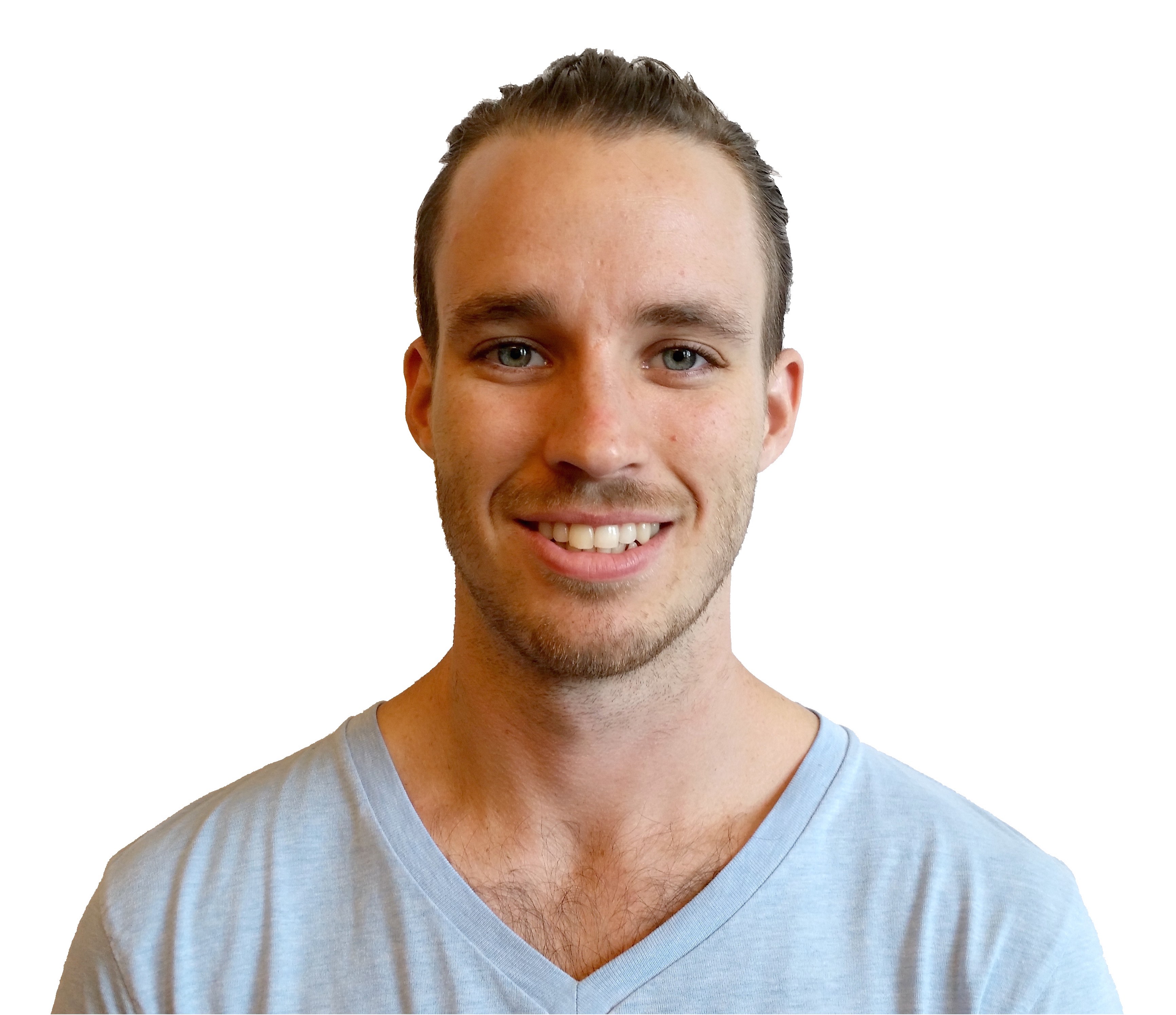}
\end{wrapfigure}
\textbf{Sean Mason} received the B.S. and M.S. degrees in mechanical
engineering from Drexel University in Philadelphia, PA in 2012. In
2015, he received the M.S. degree and is currently working towards the
Ph.D. degree in Computer Science at the University of Southern
California. His research interests include optimal control, biped
locomotion, and whole-body control frameworks for torque controlled
robots.\\

\begin{wrapfigure}[11]{l}{3.6cm}
\includegraphics[width=4cm]{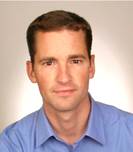}
\end{wrapfigure}
\textbf{Felix Grimminger} is a mechatronics engineer working in Stefan Schaal's Autonomous Motion Department at the Max-Planck-Institute for Intelligent Systems in T\"ubingen, Germany.
Before joining the Autonomous Motion Lab in 2013 he worked for Boston
Dynamics on the BigDog project and on several different robotic
projects at the German Research Center for Artificial Intelligence in
Bremen, Germany. He is especially interested in mechanical design,
series elastic actuation and highly dynamic legged robots.\\

\begin{wrapfigure}[10]{l}{3.6cm}
\vspace{-.75cm}
\includegraphics[width=4cm]{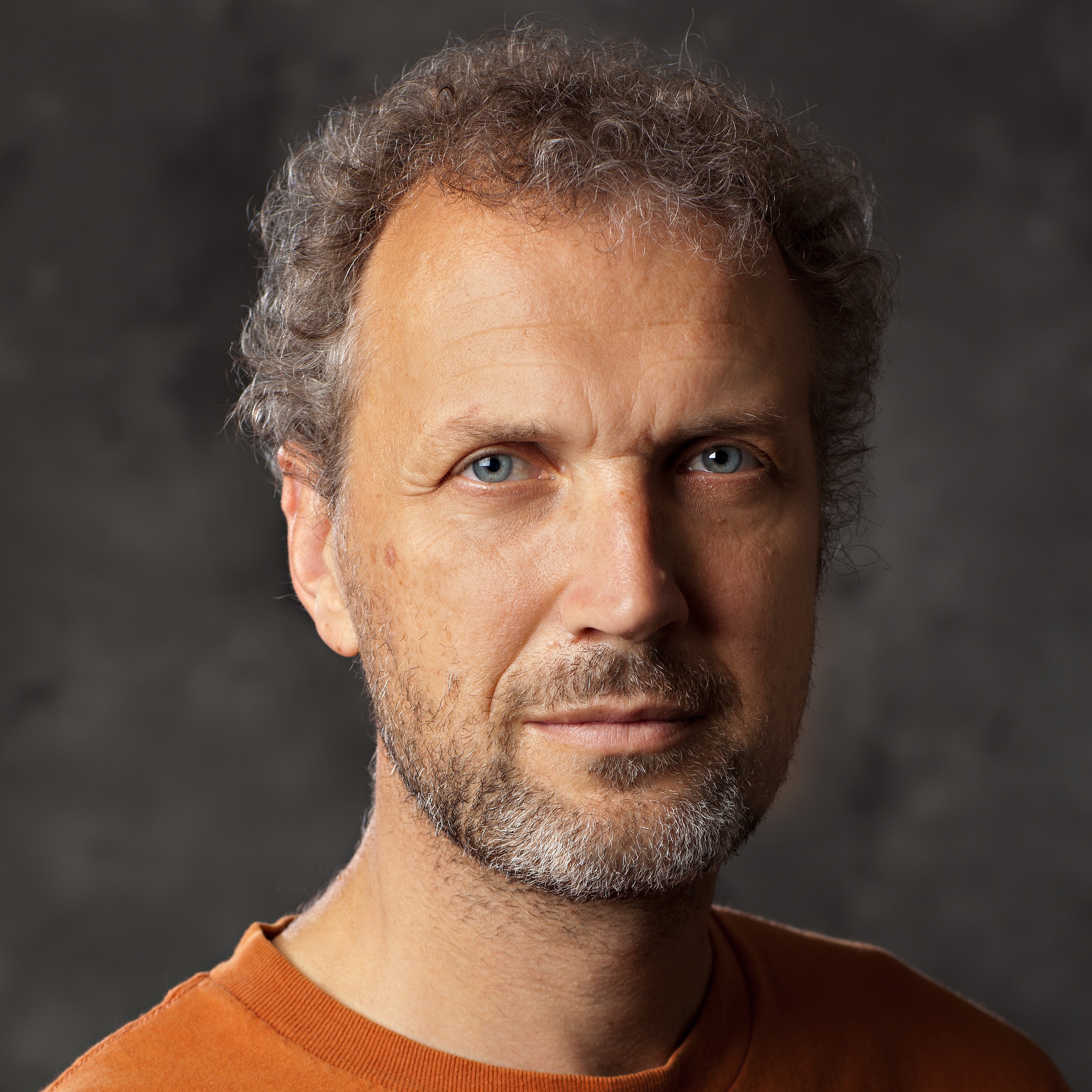}
\end{wrapfigure}
\textbf{Stefan Schaal} is Professor of Computer Science, Neuroscience, and
Biomedical Engineering at the University of Southern California, and a
Founding Director of the Max-Planck-Insitute for Intelligent Systems
in T\"ubingen, Germany. He is also an Invited Researcher at the ATR
Computational Neuroscience Laboratory in Japan, where he held an
appointment as Head of the Computational Learning Group during an
international ERATO project, the Kawato Dynamic Brain Project
(ERATO/JST).
Before joining USC, Dr. Schaal was a postdoctoral fellow at the
Department of Brain and Cognitive Sciences and the Artificial Intelligence
Laboratory at MIT, an Invited Researcher at the ATR Human Information Processing
Research Laboratories in Japan, and an Adjunct Assistant Professor at
the Georgia Institute of Technology and at the Department of Kinesiology
of the Pennsylvania State University. Stefan Schaal's research interests
include topics of statistical and machine learning, neural networks,
computational neuroscience, functional brain imaging, nonlinear dynamics, nonlinear
control theory, and biomimetic robotics. He applies his research to
problems of artificial and biological motor control and motor
learning, focusing on both theoretical investigations and experiments with human
subjects and anthropomorphic robot equipment.\\

\newpage
\begin{wrapfigure}[13]{l}{3.6cm}
\includegraphics[width=4cm]{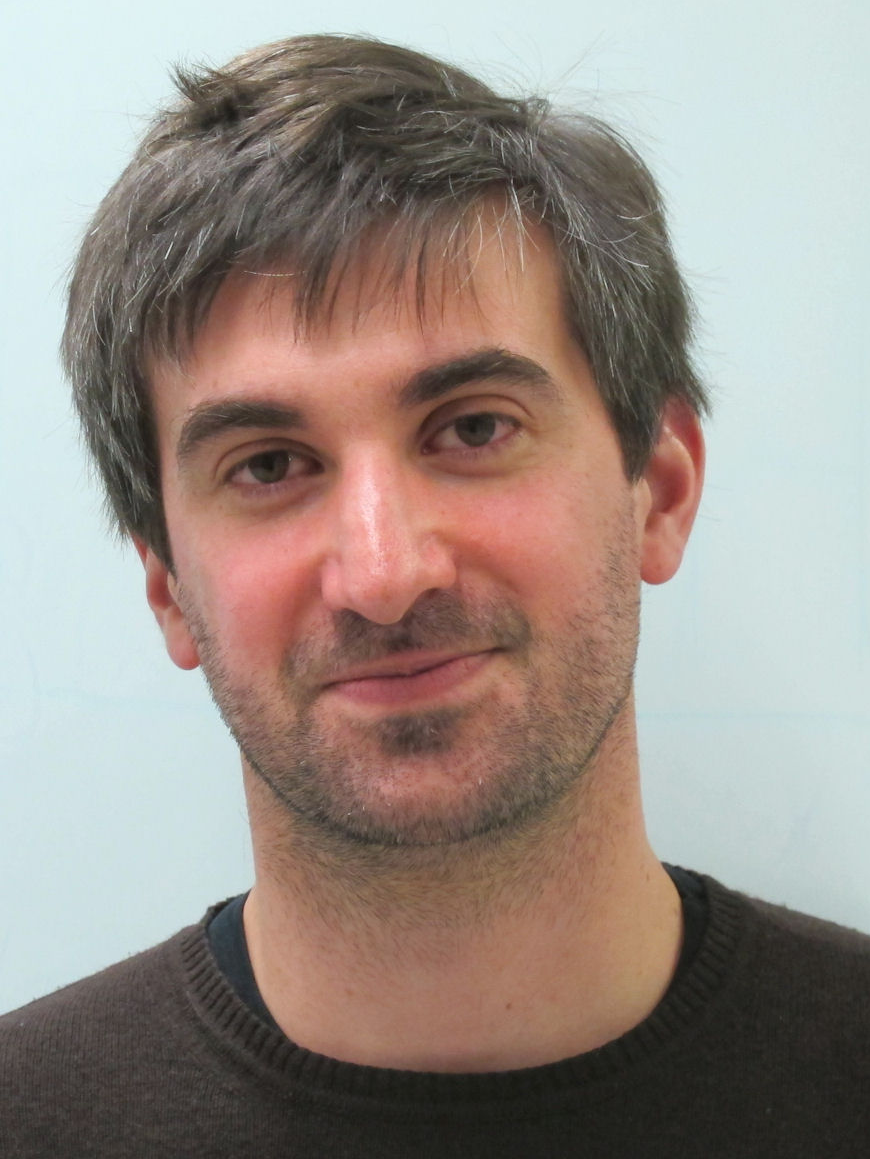}
\end{wrapfigure}
\textbf{Ludovic Righetti} leads the Movement Generation and Control group at the
Max-Planck Institute for Intelligent Systems (T\"ubingen, Germany) since
September 2012. Before, he was a postdoctoral fellow at the
Computational Learning and Motor Control Lab (University of Southern
California) between March 2009 and August 2012. He studied at the Ecole
Polytechnique F\'{e}d\'{e}rale de Lausanne where he received a diploma in
Computer Science (eq. MSc) in 2004 and a Doctorate in Science in 2008.
His research focuses on the generation and control of movements for
autonomous robots, with a special emphasis on legged locomotion and
manipulation.

\end{document}